\documentclass[runningheads]{llncs}

\usepackage{eccv}

\usepackage{eccvabbrv}

\usepackage{graphicx}
\usepackage{booktabs}
\usepackage{eccvabbrv}

\usepackage[pagebackref,breaklinks,colorlinks,citecolor=eccvblue]{hyperref}

\usepackage{array}                
\usepackage[table]{xcolor} 
\usepackage{multirow}

\usepackage{pifont}
\usepackage{adjustbox}
\usepackage{graphicx}
\usepackage{fix-cm}

\usepackage{booktabs}   
\usepackage{multirow}   
\usepackage{makecell}   

\usepackage[accsupp]{axessibility}  
\usepackage{array}                
\usepackage[table]{xcolor} 
\usepackage{multirow}
\usepackage{adjustbox}
\usepackage{graphicx}
\usepackage{fix-cm}
\usepackage{comment}
\usepackage{makecell}

\usepackage{adjustbox}
\usepackage{graphicx}
\usepackage{fix-cm}
\makeatletter
\def\thickhline{%
  \noalign{\ifnum0=`}\fi\hrule \@height \thickarrayrulewidth \futurelet
   \reserved@a\@xthickhline}
\def\@xthickhline{\ifx\reserved@a\thickhline
               \vskip\doublerulesep
               \vskip-\thickarrayrulewidth
             \fi
      \ifnum0=`{\fi}}
\makeatother
\newlength{\thickarrayrulewidth}
\usepackage{hyperref}

\usepackage{orcidlink}

\begin{document}
\newcommand{\ourmodel}{FrozenDrive}
\newcommand{\ourloss}{object-presence ratio loss}
\title{FrozenDrive: Zero-Shot Text-Guided Driving Scene Generation and Data Augmentation with Parameter-Free Frozen Diffusion Model}

\titlerunning{FrozenDrive}

\newcommand\CoAuthorMark{\footnotemark[\arabic{footnote}]}
\author{Yuhwan Jeong\orcidlink{0009-0002-0279-146X}\thanks{Equal contribution.} \and
Hyeonseong Kim\orcidlink{0009-0003-9792-4647}\protect\CoAuthorMark \and
Daehyun We\orcidlink{0009-0003-5652-1681}\protect\CoAuthorMark \and
Seonkyu Song\orcidlink{0009-0003-8282-669X}\protect\CoAuthorMark \and
Jinnyeong Yang\orcidlink{0009-0002-9275-6296}\protect\CoAuthorMark \and
Hyun-Kurl Jang\orcidlink{0009-0003-7943-3326} \and
Youngho Yoon\orcidlink{0009-0003-4346-8260} \and
Kuk-Jin Yoon\orcidlink{0000-0002-1634-2756}}

\authorrunning{Jeong et al.}


\institute{KAIST, Visual Intelligence Lab \\
\email{\{jeongyh98, brian617, greathyun, sok9854, jinnyeong6118, jhg0001, dudgh1732, kjyoon\}@kaist.ac.kr}}

\maketitle

\begin{abstract}
Synthetic data for autonomous driving is surging, powered by diffusion models that promise scalable scene generation. Yet key obstacles remain, as enforcing multi-view and temporal consistency often relies on backbone fine-tuning or added layers, which erodes pre-trained knowledge and weakens text alignment. Models also stay close to the training distribution, struggling under adverse weather and unseen configurations, and fidelity favors frequent over rare classes.
We address these gaps with \ourmodel, a controllable generative framework that preserves a pretrained diffusion model’s knowledge while achieving strong consistency. \ourmodel~conditions on rich driving-stack signals and text prompts, and introduces knowledge-preserving spatio-temporal attention to impose cross-view alignment and temporal coherence in a single pass within a parameter-free frozen diffusion backbone. An additional object-focused constraint improves per-object fidelity for rare categories.
Without any weather- or scene-specific fine-tuning, our model synthesizes globally coherent multi-view driving scenes from text, particularly under adverse and rare conditions, and surpasses prior baselines. On nuScenes, \ourmodel-augmented data significantly improves AD models performance, especially at night and in rain, demonstrating stronger robustness when trained with our scenario-targeted data.
\vspace{-5pt}

  \keywords{Multi-view generation \and Autonomous driving}
\end{abstract}

\section{Introduction}
\label{sec:intro}

Training autonomous driving models~\cite{gu2023vip3d, hu2023planning, jiang2023vad, sun2025sparsedrive, wang2024driving, yan2025rlgf} is constrained by the cost of collecting large-scale datasets. This challenge becomes even more severe when accurate labels require multi-sensor calibration and fine-grained delineation of complex scenes.
Moreover, acquiring data under adverse weather or in rare situations is intrinsically difficult: collection is error-prone, annotation is even more labor-intensive, and genuinely rare events offer few opportunities for capture, making repeated observations of the same scenario inherently difficult. As a result, real-world corpora are biased toward frequent, benign conditions, and models trained on them exhibit brittle behavior in the long tail.

Recent advances in generative modeling suggest that synthetic data can amortize annotation effort by producing diverse samples from a single label~\cite{islam2024diffusemix, li2024light, yang2023freemask}.
Mainstream methods for autonomous driving scene generation achieves 

\noindent
high visual quality under everyday conditions~\cite{wen2024panacea, gao2024magicdrive, yang2024drivearena, gao2024magicdrive3d, gao2025magicdrive, zhang2025epona, zhu2025scenecrafter, wang2025stage, guo2025dist}.
It synthesizes photorealistic driving scenes by conditioning on 2D spatial controls~\cite{zhang2023adding, zhang2020side} derived from driving-stack signals—BEV layouts, object bounding boxes, and occupancy maps~\cite{yan2025drivingsphere}. 
To encourage natural object motion, many methods reference features from the previous frame or process the entire sequence.
Regardless of the conditioning used, these methods ultimately build on the powerful generative capacity of pretrained diffusion models such as SD~\cite{rombach2022high}, SVD~\cite{blattmann2023stable}, and DiT~\cite{zheng2024open}, which are adapted to autonomous driving datasets. As a result, prior works have enabled highly controllable and realistic scene generation. Furthermore, by explicitly refining text–scene alignment during training, these methods afford fine-grained, semantically consistent editing of the scene’s weather and environment to seen conditions, \eg, transforming an originally clear setting into rain or night conditions observed in training dataset.

\begin{figure}[t]
    \centering
    \includegraphics[width=.99\linewidth]{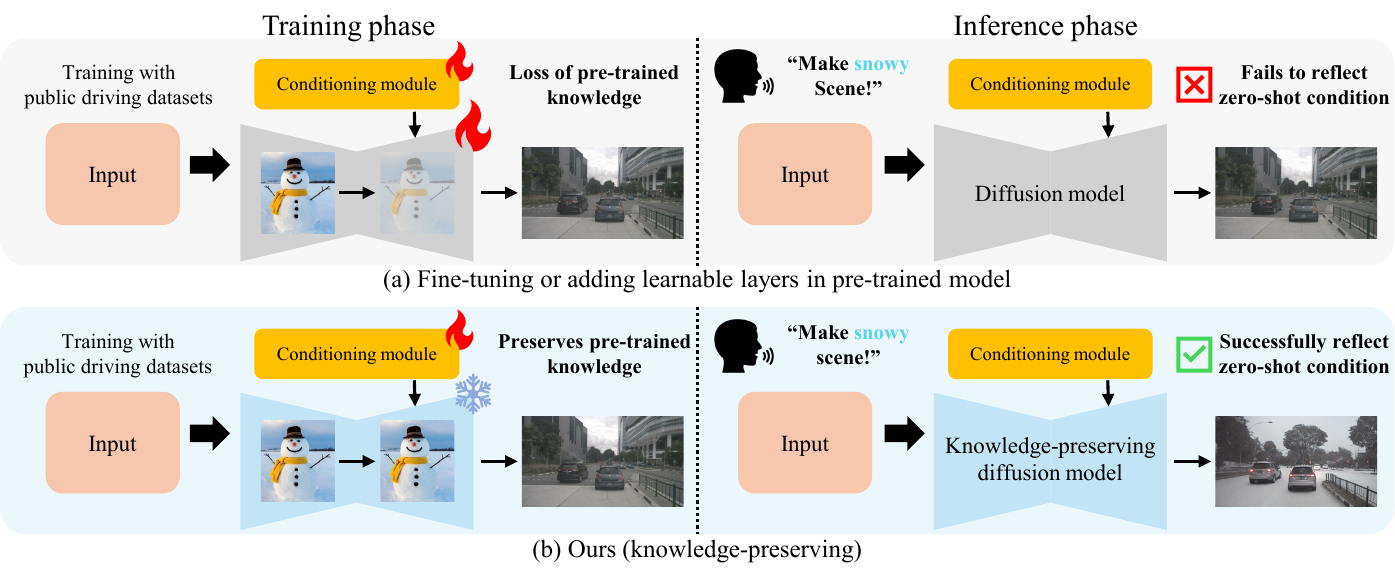}
    \vspace{-5pt}
    \caption{(a) Fine-tuning or adding learnable layers lose pre-trained priors and fails to reflect unseen text prompts while training the model.
    (b) By completely freezing the pretrained model without introducing any additional parameters, our method preserves its rich generative priors, enabling successful zero-shot text-guided generation.}
    \label{fig:teaser}
    \vspace{-12pt}
\end{figure}

These works, in order to support multi-camera rigs and temporal dynamics, either (i) generate each view largely independently with weak feature sharing~\cite{swerdlow2024street, zhu2025scenecrafter}, or (ii) introduce cross-view and temporal attention via additional layers fine-tuned on top of a pretrained backbone~\cite{wen2024panacea, gao2024magicdrive, yang2024drivearena, yan2025drivingsphere}.

While these designs improve consistency under seen conditions, they remain heavily bounded by the training distribution. Models trained on major benchmarks~\cite{caesar2020nuscenes, sun2020scalability} perform well in standard daytime scenarios but suffer severe degradation in underrepresented settings. This degradation primarily stems from extensive backbone fine-tuning on datasets with significant coverage gaps. Consequently, the rich priors of the original pretrained model for adverse or unseen scenarios (\eg, heavy snowfall) are largely overwritten or catastrophically forgotten. Furthermore, such reliance on full backbone updates not only compromises the pretrained text-to-image alignment but also fails to preserve per-object fidelity, particularly for rare classes.

To address these limitations, we propose \ourmodel, a generative framework that preserves a pretrained diffusion model’s knowledge while enabling precise control over multi-view imagery under adverse or unseen scenarios.
Consistency is enforced via two knowledge-preserving plug-in attention mechanisms, multi-view inflated self-attention and temporal reference self-attention, that operate jointly within the attention pathway: features from all camera views are concatenated to impose circular cross-view alignment, while features from preceding frames are injected to endow video-style temporal coherence, all in a single pass.
\textbf{Yet the most consequential difference lies elsewhere: we neither introduce additional parameters nor update the pretrained weights.} 
By keeping the diffusion backbone frozen and parameter-free, \ourmodel~preserves the pretrained text–image alignment and avoids knowledge erosion, thereby retaining zero-shot text guidance.
In contrast, prior approaches~\cite{wen2024panacea, gao2024magicdrive, yang2024drivearena, yan2025drivingsphere} typically strengthen consistency by tuning the backbone, which can weaken alignment and lead to local or superficial edits. As shown in~\cref{fig:teaser}, such methods (\eg, DriveArena~\cite{yang2024drivearena}) often struggle to compose plausible snow scenes unseen during training, whereas \ourmodel~produces globally coherent results without any weather-specific fine-tuning.

To further secure per-object fidelity, we impose an object-focused constraint for rare objects. This shifts learning toward underrepresented categories, improving their generation quality rather than privileging only frequent classes.
Together, our design secures robust text alignment and multi-camera consistency, yielding diverse, high-quality street views. 
With these components, our synthetic data captures fine-grained nuScenes details and achieves state-of-the-art perception and planning performance evaluated with UniAD~\cite{hu2023planning}.

Moreover, we leverage text prompts to specify a wide range of driving scenarios, from common conditions to rare, hard-to-collect cases.
Consequently, \textbf{\ourmodel~enables scenario-targeted data generation that strengthens downstream robustness}: 
AD models trained with \ourmodel-augm- ented data achieve superior perception and planning performance under adverse conditions (\eg, night and rain) on nuScenes, outperforming prior methods.

Our main contributions can be summarized as follows:
\begin{itemize}
\item We introduce \ourmodel, a knowledge-preserving driving scene generator that enforces multi-view and temporal consistency on a parameter-free frozen pretrained diffusion backbone, enhanced with object-balanced constraints and structured conditioning signals.
\item We enable text-guided compositional control over diverse scene factors, time of day, weather conditions, and rare environments, supporting zero-shot synthesis of unseen combinations.
\item We demonstrate improved synthesis quality and superior transfer to downstream autonomous driving tasks under both common and adverse weather.
\end{itemize}


\section{Related works}
\label{sec:rel}

\subsection{Diffusion models}

Diffusion models~\cite{ho2020denoising, song2021scorebased, song2020denoising, ho2022classifier} synthesize images via iterative denoising but are memory-prohibitive at high resolutions in pixel space. Latent diffusion models~\cite{rombach2022high} address this by operating in a latent space, enabling efficient, high-fidelity generation and large-scale text-conditioned pretraining further broadens their applications. Recent extensions to video explicitly model temporal consistency for coherent synthesis~\cite{zheng2024open, blattmann2023stable, he2022latent, yang2025cogvideox, wang2025wan}. In autonomous driving, video diffusion models~\cite{guo2025dist, wen2024panacea, hu2023gaia, russell2025gaia, lu2024wovogen, zhao2025drivedreamer, lu2025infinicube} likewise target temporal coherence, but often add extra conditioning layers that increase memory and weaken pretrained priors.

\subsection{Multi-view generation in autonomous driving}
Advances in multi-view driving scene generation~\cite{zhou2025hermes, ji2025cogen, mei2024dreamforge, zhu2025consistentcity, wang2025authentic, yang2025xscene, yang2026consisdrive} have improved the realism and controllability of autonomous-driving simulators. For multi-view generation~\cite{lu2024seeing, jiang2025dive, dong2025noisecontroller, li2024drivingdiffusion}, systems commonly inject structured cues, 3D bounding boxes, BEV maps, ego trajectories, and multi-camera poses, via ControlNet to enforce spatial and geometric constraints across views~\cite{gao2024magicdrive, gao2024magicdrive3d, yang2024drivearena, li2025dualdiffdualbranchdiffusionmodel, wu2024drivescape, zhang2025epona, swerdlow2024street, yang2025instadrive}. BEV-based frameworks~\cite{swerdlow2024street,yang2023bevcontrol, liu2024mvpbev} enhance scene-level spatial coherence. Panacea~\cite{wen2024panacea} uses BEV and 4D spatiotemporal attention for coherence. MagicDrive~\cite{gao2024magicdrive} adopts tailored encoders and cross-view attention for precise 3D geometry. DriveArena~\cite{yang2024drivearena} uses a reference frame as a conditioning signal to propagate information across sequences.
Moreover, MagicDrive-V2~\cite{gao2025magicdrive}, Genesis~\cite{guo2025genesis} and DrivingSphere~\cite{yan2025drivingsphere} use DiT architectures, enabling high-resolution 3D scene reconstruction. DiST-4D~\cite{guo2025dist} leverages metric depth to align spatial structure and temporal continuity.
These methods typically rely on extra training or parameters to enforce multi-view and temporal consistency.



\subsection{Scene-level data augmentation}
Scene-level data augmentation~\cite{abu2018augmented, chen2021geosim, xiao2022polarmix, ryu2023instant, yang2020small, li2022traffic}, enhances data diversity by editing or transforming existing scenes, thereby complementing costly real-world data collection.
One major approach is spatial manipulation~\cite{tong20243d, cheng2025object, li2023climatenerf}, which involves reconstructing environments in 3D using NeRF~\cite{mildenhall2021nerf, li2023read, feldmann2024nerfmentation} or Gaussian Splatting~\cite{kerbl20233d, yang2025novel, zhao2025drivedreamer4d, ni2025recondreamer, zhu2026worldsplat}, and then re-rendering them to preserve multi-view geometric consistency. Meanwhile, recent works~\cite{zhu2025scenecrafter, liang2025driveeditor, zhou2024simgen, wu2024improving} leverage diffusion-based scene editing to achieve geometry-aware manipulations.
Another major approach is style-based augmentation~\cite{islam2024diffusemix, li2024light}, which modifies the visual domain without changing the scene geometry. These methods, implemented via neural style transfer, can be categorized into three strands: (i) condition-specific transfer adjusts weather, lighting, and season to narrow out-of-distribution gaps~\cite{li2021weather, assion2024bdd, 10960638, 10870179, rothmeier2024time, zhang2022simbar}; (ii) semantic-guided synthesis uses pixel-aligned masks or layouts for high-fidelity edits while preserving labels~\cite{wang2022semantic, zeng2023scenecomposer, zhang2023adding}; (iii) instruction-driven editing~\cite{gatys2016image, brooks2022instructpix2pix, hertz2022prompt, meng2022sdedit} follows natural-language prompts, often via cross-attention control. Building on these advances, we focus on multi-view driving scene generation and propose a framework that synthesizes diverse, geometrically consistent driving scenarios from a text prompt.


\section{Methods}
\label{sec:methods}

\subsection{Preliminary: latent diffusion models}
Latent Diffusion Models (LDM)~\cite{rombach2022high} train the diffusion process in a learned latent space. A perceptual autoencoder $(E,D)$ maps images to latents $\mathbf{z}=E(\mathbf{x})$ and reconstructs $\hat{\mathbf{x}}=D(\mathbf{z})$, reducing spatial complexity while preserving semantics. Diffusion is applied to $\mathbf{z}$, \ie, $q(\mathbf{z}_t|\mathbf{z}_0)$ and $p_\theta(\mathbf{z}_{t-1}|\mathbf{z}_t)$, with the noise-prediction loss. Conditioning (\eg, text) is injected via cross-attention inside the U-Net backbone, enabling flexible, scalable generation with substantially reduced memory and compute compared to pixel-space diffusion. In \ourmodel, we adopt the pretrained LDM, Stable Diffusion, for multi-view image generation, as it is lightweight, highly adaptable, and widely used~\cite{gao2024magicdrive,wen2024panacea,yang2024drivearena,yang2025xscene}. While others customize the SD, we use it frozen to leverage all the knowledge it contains.

\subsection{Key insight: knowledge preservation}
Pretrained LDMs, when used directly, are difficult to guide toward reliable generation of a realistic multi-view driving scene.
A common remedy is to adapt the model with driving scene data—by (1) fine-tuning all or part of the backbone, (2) adding trainable adapters or modules, or (3) using a ControlNet-based conditioning pipeline~\cite{zhang2023adding}; these strategies are often combined. Our goal, however, is not only to synthesize plausible scenes but to retain the pretrained prior so that text-guided \emph{zero-shot} synthesis remains effective.



Guided by this goal, we consider these strategies through the lens of knowledge preservation. It is widely recognized that full or partial fine-tuning~\cite{he2021analyzing, park2024textboost, zhao2024continual, liao2025continual, ming2024does, kotha2024understanding, li2022domain, kumari2023multi} can fit the training set while eroding the pretrained prior, weakening text alignment and zero-shot ability. More critically, we empirically observe that adding only a few trainable parameters (\eg, multi-view or temporal cross-attention attached to the backbone) induces similar drift, narrowing visual diversity and weakening prompt fidelity (see Sec.~\ref{sec:forgetting}).
%
%
%
These observations motivate a simple principle: \emph{freeze the backbone to preserve knowledge}. We therefore adopt ControlNet-based conditioning, where the LDM remains intact while ControlNet ingests structured signals and steers generation toward driving scenes. This design preserves the model’s text understanding and enables robust zero-shot synthesis, while still allowing precise, task-specific control.

\begin{figure}[t]
    \centering
    \vspace{-2pt}
    \includegraphics[width=.98\textwidth]{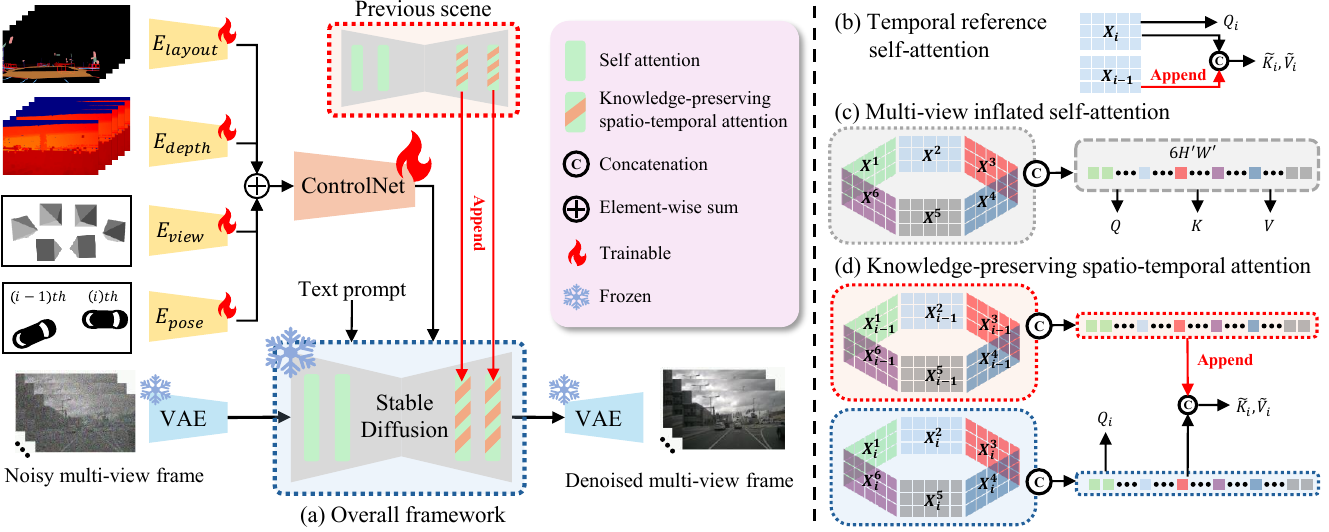}
    \vspace{-8pt}
    \caption{\textbf{Overall framework.} (a) \ourmodel~uses five conditions: (i) an HD map, (ii) per-view camera indicator, (iii) depth map, (iv) relative pose, and (v) text description. The first four are multi-view, pixel- and spatially aligned and injected into the diffusion model via ControlNet, while the last enables zero-shot text guidance.
    After the encoder, we replace original self-attention into our knowledge-preserving spatio-temporal attention. (b) Temporal reference self-attention enforces frame-to-frame consistency by expanding the input context to reuse the previous frame as a reference. (c) Multi-view inflated self-attention enforces cross-view consistency by concatenating latent features from all camera views. (d) Knowledge-preserving spatio-temporal attention reshapes self-attention so that multi-view inflated and temporal reference self-attention act jointly. During this process, the projection layers are kept frozen.}
    \vspace{-10pt}
    \label{fig:architecture}
\end{figure}

\subsection{Overall pipeline}
\Cref{fig:architecture} provides an overview of our pipeline. For synthesis, we use (i) an HD map with road-layout layers and 3D bounding boxes, (ii) per-view camera indicator, (iii) depth derived from an occupancy map, (iv) the relative pose to the previous frame, and (v) a text description. Each non-text signal $c_k$ is passed through a lightweight embedding network $E_k$ to produce an embedding $\mathbf{e}_k = E_k(c_k)$ in the latent space. These embeddings $\{\mathbf{e}_{\text{layout}}, \mathbf{e}_{\text{view}}, \mathbf{e}_{\text{depth}}, \mathbf{e}_{\text{pose}}\}$ are added to the latent feature map $\mathbf{z}$ and jointly serve as input to ControlNet.
The resulting control features 
modulate the frozen diffusion backbone, while text is encoded separately by the frozen text encoder. Together, they guide generation without modifying or adding parameters to the diffusion backbone.
Details of the embedding design are provided in the \textit{supplementary materials}.
%
%
To preserve prior knowledge, we keep this backbone frozen and enforce multi-view/temporal consistency by modifying only the self-attentions, \ie, which features are attended(\cref{sec:mv}), integrating rich spatial cues while avoiding pretrained knowledge forgetting. Finally, because fidelity tends to track observation frequency, the long-tail distribution impairs rare-class rendering, and we mitigate this with an object-focused ratio loss that upweights underrepresented classes (\cref{sec:obloss}).

\subsection{Knowledge-preserving multi-view generation}
\label{sec:mv}
Our objective is to synthesize realistic autonomous driving scenes across multiple cameras and over time while preserving the prior of a powerful pretrained diffusion model.
We therefore adopt a purely ControlNet-based pipeline.
%
%
Yet directly conditioning a frozen generator is insufficient for our setting: multi-camera perception demands \textbf{cross-view consistency} (overlapping views must agree on geometry, appearance, and semantics), and video rollout requires \textbf{temporal consistency} (successive frames must remain coherent under ego- and object motion without hallucinated discontinuities). Enforcing both under a knowledge-preserving regime is challenging, as we introduce no new trainable parameters into the diffusion backbone.

To address this, we propose a \emph{knowledge-preserving spatio-temporal attention} that operates on the frozen LDM. Rather than learning new attention modules or modifying existing parameters, we reinterpret how the original self-attention layers are \emph{fed} by manipulating their inputs. The block is composed of two complementary mechanisms: (i) \emph{multi-view inflated self-attention}, which promotes cross-view agreement by allowing features from different camera views to attend to one another and (ii) \emph{temporal reference self-attention}, which encourages frame-to-frame stability by letting the current frame attend to the previous frame as an explicit reference. Both mechanisms reshape the attention context seen by each layer while leveraging conditional signals injected by ControlNet, leaving the attention weights themselves untouched. We next describe the attention module and its auxiliary embeddings.

\noindent
\textbf{Multi-view inflated self-attention.}
To enforce cross-view consistency while preserving the pretrained prior, we introduce multi-view inflated self-attention (MISA). MISA concatenates latent features from all $n_{\text{view}}$ cameras and performs a single self-attention pass with the frozen layer, enabling cross-view exchange beyond standard within-view attention. ControlNet supplies per-view conditioning features, while MISA expands the attention field across views. 
Let $\mathbf{X}^{(v)} \in \mathbb{R}^{N_v \times d_{\text{model}}}$ denote the latent feature token for view $v$, where $N_v$ is the number of tokens for view $v$ and $d_{\text{model}}$ is the token embedding width. We form a joint sequence and apply shared self-attention:
\begin{equation}
\tilde{\mathbf{X}} := \big[\, \mathbf{X}^{(1)} \,;\, \mathbf{X}^{(2)} \,;\, \cdots \,;\, \mathbf{X}^{(n_{\text{view}})} \,\big],
\end{equation}
\vspace{-11pt}
\begin{equation}
\mathrm{Attn}(\tilde{\mathbf{X}})\!
=\!
\mathrm{softmax}\!\left(\!
\frac{(\tilde{\mathbf{X}}\mathbf{W}^Q)(\tilde{\mathbf{X}}\mathbf{W}^K)^{\top}}{\sqrt{d_h}}\!
\right)\!
(\tilde{\mathbf{X}}\mathbf{W}^V),
\end{equation}
where $\mathbf{W}^Q,\mathbf{W}^K,\mathbf{W}^V \in \mathbb{R}^{d_{\text{model}}\times d_{\text{model}}}$ are frozen LDM parameters and $d_h$ is the per-head dimension.


We also indicate which view produced each feature via a lightweight view embedding $\mathbf{e}_{\text{view}}$. Each view index $v\!\in\!\{1,\dots,n_{\text{view}}\}$ is encoded with Fourier features~\cite{tancik2020fourier} and supplied to ControlNet as an additional conditioning input.
%
%
This simple tag provides view identity and neighborhood structure, improving cross-view correspondence without introducing any new attention parameters.


\noindent
\textbf{Temporal reference self-attention.}
To enforce frame-to-frame consistency under knowledge preservation, we introduce temporal reference self-attention (TRSA), which, analogously to MISA, keeps all LDM attention weights frozen and instead expands their input context to reuse the previous frame as a reference. 
Concretely, for the current frame $i$ and a reference frame $i-j$, we take their latent features at the same noise level, $\mathbf{X}_i$ and $\mathbf{X}_{i-j}$, and compute the projections using frozen weights. We obtain the query, key, and value for the current frame as $\mathbf{Q}_i=\mathbf{X}_i\mathbf{W}^Q$, $\mathbf{K}_i=\mathbf{X}_i\mathbf{W}^K$, and $\mathbf{V}_i=\mathbf{X}_i\mathbf{W}^V$, and the key and value for the reference frame as $\mathbf{K}_{i-j}=\mathbf{X}_{i-j}\mathbf{W}^K$ and $\mathbf{V}_{i-j}=\mathbf{X}_{i-j}\mathbf{W}^V$.

We then augment the key/value banks of the current frame with the reference, and the self-attention is computed as :
\begin{equation}
\tilde{\mathbf{K}}_i=\big[\mathbf{K}_i\,;\,\mathbf{K}_{i-j}\big],\quad
\tilde{\mathbf{V}}_i=\big[\mathbf{V}_i\,;\,\mathbf{V}_{i-j}\big], 
\end{equation}
\vspace{-11pt}
\begin{equation}
\mathrm{Attn}(\mathbf{X}_i)=
\mathrm{softmax}\!\left(\frac{\mathbf{Q}_i \tilde{\mathbf{K}}_i^{\top}}{\sqrt{d_h}}\right)\tilde{\mathbf{V}}_i.
\end{equation}
%
Exposing each current features to keys/values from the previous frame enables the retrieval of temporally aligned appearance and geometry cues (\eg, object, weather), stabilizing the rollout without introducing new parameters.


We also encode inter-frame motion with a lightweight relative pose embedding $\mathbf{e}_{\text{pose}}$ between frames $i{-}j$ and $i$. Instead of passing the raw 6-DoF transform, we construct a spatial map: for each image location $(x,y)$, lift to $(x,y,0)$, transform by the known 3D relative pose, take the in-plane coordinates $(x',y')$, and encode them with Fourier features~\cite{tancik2020fourier}. This pose-aware map is injected via ControlNet as additional conditioning, providing dense correspondence cues without introducing new attention parameters.

\begin{figure}[t]
    \centering
    \includegraphics[width=.98\linewidth]{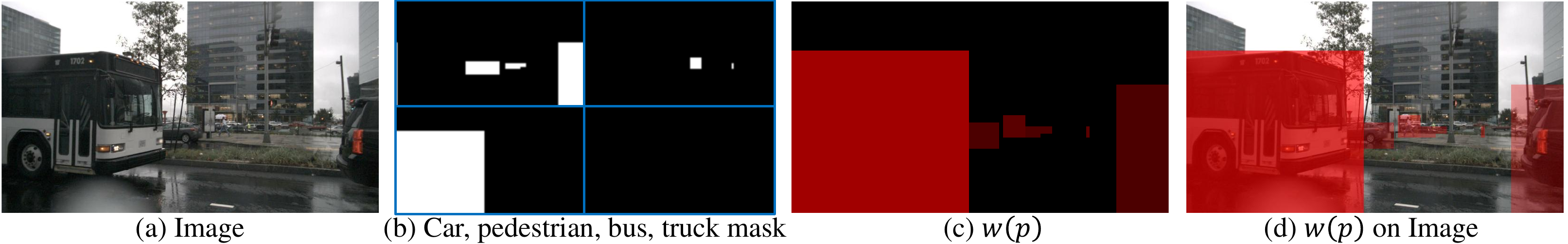}
    \vspace{-7pt}
    \caption{\textbf{Details of the class-wise weighting mask.} (a) Input image. (b) Binary masks for car, pedestrian, bus, and truck. (c) Weighted mask $w(p)$ obtained by scaling each category by its object-presence ratio. In overlapping regions, we apply a max rule and keep the larger value. (d) Overlay of $w(p)$ on the image.}
    \label{fig:loss}
    \vspace{-13pt}
\end{figure}

\noindent
\textbf{Knowledge-preserving spatio-temporal attention.}
%
Finally, we unify the above mechanisms by reshaping self-attention so that MISA and TRSA act jointly within a single, frozen block. For effective interaction with ControlNet, we apply this spatio-temporal attention only in the LDM decoder: ControlNet supplies scene-level conditioning (\eg, HD maps, depth, view, and relative poses), and the decoder fuses these signals with our attention to produce view-consistent, temporally stable generations. 
This design keeps the diffusion backbone frozen and parameter-free, thereby preserving the model's generalization capability.





\subsection{Objective function}
\label{sec:obloss}


\noindent
Guided by our design, we train ControlNet and the embedders, keeping the diffusion backbone frozen, to predict noise with the standard DDPM~\cite{ho2020denoising} objective:

\begin{equation}
\mathcal{L}
= \mathbb{E}_{x_{0},c,\epsilon,t}
\bigl[
\bigl\|
\epsilon - \epsilon_{\theta}\!\bigl(
\sqrt{\bar{\alpha}_t}\,x_0 + \sqrt{1-\bar{\alpha}_t}\,\epsilon,\; t,\; c
\bigr)
\bigr\|^{2}
\bigr].
\end{equation}

Supervised driving data are often long-tailed (\eg, in the nuScenes dataset, \emph{bicycle} appears at only $\sim$2.3\% of the \emph{car} frequency; Table~{A}, \textit{supple.}), which hurts fidelity for rare classes. To mitigate this, we propose \ourloss, an object-balanced ratio loss that emphasizes underrepresented categories via a per-pixel weight map derived from projected 3D boxes (see~\cref{fig:loss}):
\vspace{-2pt}
\begin{gather}
\mathcal{L}_{\text{total}}
=\frac{1}{|\Omega|}\sum_{p\in\Omega}\bigl(1+\lambda\,w(p)\bigr)\,\mathcal{L}(p),\\
w(p)=\max_{k\in\mathcal{K}_p} w_k,\quad
\mathcal{K}_p=\{\,k\mid p\in\mathcal{M}_k\,\}.
{\vspace{-4pt}}
\end{gather} 
Here, $\Omega$ is the image lattice and $\mathcal{L}(p)$ the per-pixel diffusion loss. $\mathcal{M}_k$ denotes the foreground mask obtained by projecting class-$k$ 3D boxes and $w_k$ is a class-dependent weight (larger for rarer classes). When there is no object mask at pixel $p$, \ie, $\mathcal{K}_p=\varnothing$, set $w(p)=0$. The max-overlap rule lets the rarest overlapping instance dominate the weighting.

\begin{table}[t]
\centering
\caption{\textbf{Generation fidelity experiments.} We generate multi-view images under nuScenes~\cite{caesar2020nuscenes} validation set conditions. Metrics are computed by evaluating the synthesized results with UniAD~\cite{hu2023planning} and measuring FVD~\cite{unterthiner2018towards}. \textbf{Bold} and \underline{underline} denote the best and second-best, respectively.}
\vspace{-5pt}
\resizebox{1.0\textwidth}{!}{
\setlength{\tabcolsep}{3.0pt}
\begin{tabular}{l|c|c|cc|cccc|ccc>{\columncolor{gray!20}}c|c}
    \thickhline
    \multicolumn{1}{c|}{\multirow{2}{*}{Method}} &
    \multicolumn{1}{c|}{\multirow{2}{*}{Venue}} &
    \multicolumn{1}{c|}{\multirow{2}{*}{Backbone}} &
    \multicolumn{2}{c|}{3DOD ($\uparrow$)} &
    \multicolumn{4}{c|}{BEV Segmentation mIoU ($\uparrow$)} &
    \multicolumn{4}{c|}{L2 ($\downarrow$)} &
    \multicolumn{1}{c}{Quality ($\downarrow$) } \\ \cline{4-14}
    & &  &  mAP & NDS & Lanes & Drivable & Divider & Crossing & 1.0s & 2.0s & 3.0s & Avg. & FVD \\ \thickhline 
    nuScenes~\cite{caesar2020nuscenes} & - & - & 37.98 & 49.85 & 31.31 & 69.14 & 25.93 & 14.36 & 0.51 & 0.98 & 1.65 & 1.05 & - \\ \hline
    MagicDrive~\cite{gao2024magicdrive} & ICLR'24 & SD~\cite{rombach2022high} &  12.92 & 28.36 & 21.95 & 51.46 & 17.10 & 5.25 & 0.57 & 1.14 & 1.95 & 1.22 & 218.1 \\
    Panacea~\cite{wen2024panacea} & CVPR'24 &  SD~\cite{rombach2022high} & 13.72 & 27.73 & 18.23 & 52.37 & 17.21 & - & 0.58 & 1.14 & 1.97 & 1.23 & 139.0  \\
    DrivingSphere~\cite{yan2025drivingsphere} & CVPR'25 & STDiT~\cite{zheng2024open} & \underline{21.45} & \underline{34.16} & 27.99 & \underline{62.87} & 22.29 & - & 0.54 & 1.10 & \underline{1.76} & 1.13 & 103.4 \\
    DiST-4D~\cite{guo2025dist} & ICCV'25 & STDiT~\cite{zheng2024open}& 15.63 & 32.44 & 26.80 & 60.32 & 21.69 & \underline{10.99} & 0.56 & 1.11 & 1.91 & 1.19  & \textbf{22.6} \\
    DriveArena~\cite{yang2024drivearena} & ICCV'25 & SD~\cite{rombach2022high} &  16.06 & 30.03 & 26.14 & 59.37 & 20.79 & 8.92 & 0.56 & 1.10 & 1.89 & 1.18 & 185.3 \\
    MagicDrive-V2~\cite{gao2025magicdrive} & ICCV'25 & STDiT~\cite{zheng2024open} & 15.24 & 31.25 & 25.02 & 58.63 & 19.84 & 9.98 & \textbf{0.49} & \underline{1.00} & 1.78 & \underline{1.09}  & \underline{81.6} \\
    X-Scene~\cite{yang2025xscene} & NeurIPS'25 & SD~\cite{rombach2022high} & 20.40 & 31.76 & \underline{28.04} & 61.96 & \underline{22.32} & 10.48 & 0.55 & 1.08 & 1.81 & 1.15 & 179.7 \\
    \textbf{FrozenDrive (Ours)} & - & SD~\cite{rombach2022high} & \textbf{21.87} & \textbf{35.32} & \textbf{28.88} & \textbf{64.27} & \textbf{23.58} & \textbf{11.66} & \underline{0.50} & \textbf{0.98} & \textbf{1.66} & \textbf{1.05}  & 136.8 \\ \thickhline
\end{tabular}}
\vspace{-5pt}
\label{tab:main}
\end{table}

\begin{figure}[t]
\centering
\includegraphics[width=.99\textwidth]{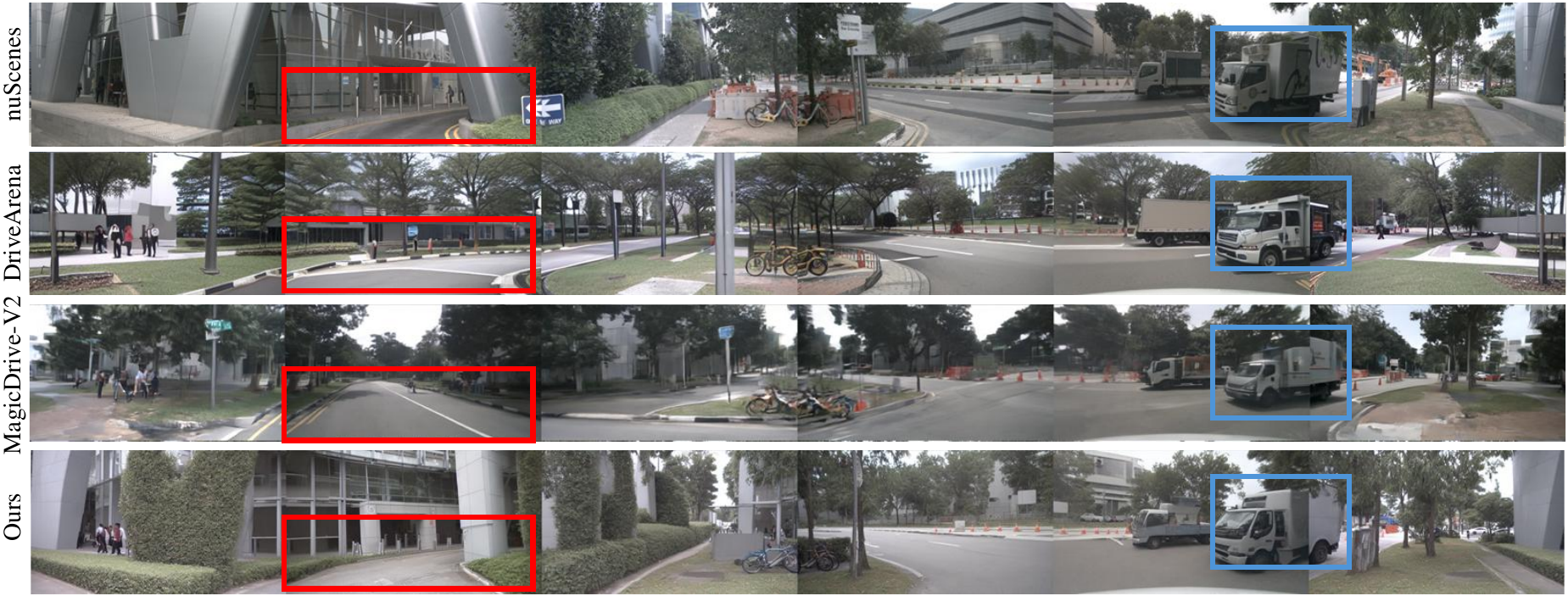}
\vspace{-5pt}
\caption{\textbf{Qualitative comparison of generated samples} under nuScenes~\cite{caesar2020nuscenes} validation conditions. Using the same nuScenes conditions, we generate multi-view images with each method to evaluate visual quality. Compared to the ground truth (top row), MagicDrive-V2~\cite{gao2025magicdrive} produces structural artifacts in road conditions (red boxes), and DriveArena~\cite{yang2024drivearena} exhibits view-inconsistent object shapes (blue boxes). Our method better preserves scene layout while maintaining superior multi-view consistency.}
\label{fig:qual}
\end{figure}

\begin{figure}[t]
\centering
\includegraphics[width=.99\linewidth]{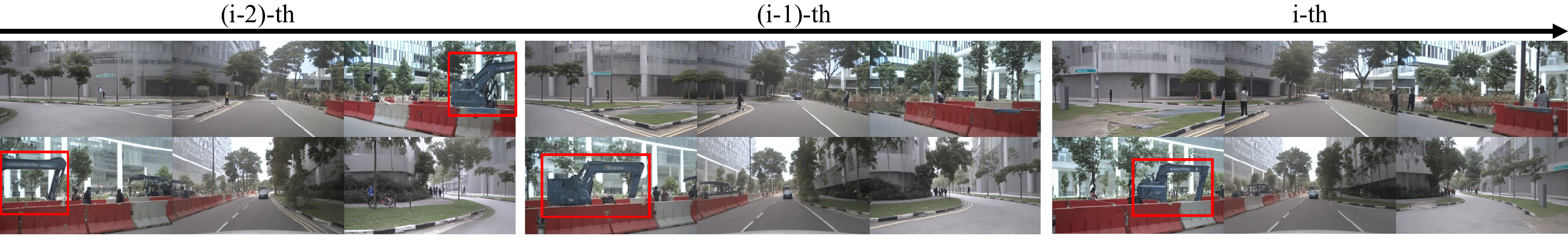}
\vspace{-5pt}
\caption{\textbf{Visualization of sequential generation.} \ourmodel~produces high-fidelity, temporally coherent images across scenes. The rare construction vehicle (red boxes) is faithfully preserved in shape, color, and position across viewpoints and consecutive frames, demonstrating strong object-level consistency.}
\label{fig:qual_temporal}
\end{figure}


\section{Experiments}
\label{sec:exp}
\subsection{Experimental settings} We train and evaluate our proposed method on the nuScenes~\cite{caesar2020nuscenes} dataset. It provides rich sensor data and annotations, including LiDAR point clouds, multi-view images, lane information, poses, and related labels. Similar to prior work~\cite{gao2024magicdrive, wang2023we}, the 2Hz keyframe annotations are temporally interpolated to obtain 12Hz labels. For depth map, we stack LiDAR point cloud to generate occupancy and measure the depth to ego position~\cite{li2025uniscene}.
We first assess the \ourmodel~generation quality in Sec.~\ref{sec:gen}. In Sec.~\ref {sec:adv}, we further evaluate the impact of \ourmodel-based data augmentation on adverse conditions and measure how these samples improve the generalization of an existing AD model.


\subsection{Generation fidelity}
\label{sec:gen}
%
\noindent\textbf{Metric.} Following previous studies~\cite{yang2024drivearena, yan2025drivingsphere, gao2024magicdrive}, we compare the performance of \ourmodel~with other methods in terms of perception and planning via UniAD~\cite{hu2023planning}. For generation quality, we utilize FVD~\cite{unterthiner2018towards} to evaluate.

\noindent\textbf{Baselines} are MagicDrive~\cite{gao2024magicdrive}, Panacea~\cite{wen2024panacea}, DrivingSphere~\cite{yan2025drivingsphere}, DiST-4D~\cite{guo2025dist}, DriveArena~\cite{yang2024drivearena}, MagicDrive-V2~\cite{gao2025magicdrive}, and X-Scene~\cite{yang2025xscene} which generates multi-view images under nuScenes dataset conditions.

\begin{table}[t!]
\centering                             
\caption{\textbf{Impact of Synthetic Night/Rain Data on nuScenes Perception and Planning.}
Perception and planning performance of SparseDrive~\cite{sun2025sparsedrive} on nuScenes under adverse conditions (night and rain) with different data generation strategies. The baseline uses only original normal-weather samples, while rule-based methods~\cite{li2019heavy,thompson2002spatial} translate normal samples into night/rain images with hand-crafted filters that introduce artifacts. DriveArena~\cite{yang2024drivearena}, MagicDrive-V2~\cite{gao2025magicdrive}, and \ourmodel~synthesize condition-aware data from text prompts. \textbf{Bold} and \underline{underline} denote the best and second-best.
}
\vspace{-5pt}
\resizebox{1.0\textwidth}{!}{
\setlength{\tabcolsep}{6.8pt} 
\begin{tabular}{l|l|cc|ccc>{\columncolor{gray!20}}c|ccc>{\columncolor{gray!20}}c}
\thickhline
\multirow{2}{*}{} &
  \multicolumn{1}{c|}{\multirow{2}{*}{Method}} &
  \multicolumn{2}{c|}{3DOD ($\uparrow$)} &
  \multicolumn{4}{c|}{Online mapping results ($\uparrow$)} &
  \multicolumn{4}{c}{L2(m) ($\downarrow$)}  \\ \cline{3-12}
 &  & mAP & NDS & $AP_{ped}$ & $AP_{divider}$ & $AP_{boundary}$ & mAP & 1.0s & 2.0s & 3.0s & Avg. \\ \thickhline 
\multirow{5}{*}{\rotatebox{90}{Night}} & Baseline &6.62 & 15.19 & 0.07 & 8.48 & 9.42 & 5.99 & 0.74 & 1.36 & 2.11 & 1.40 \\
& Rule-based~\cite{thompson2002spatial} & 7.42 & 13.11 & 0.30 & 8.93 & 10.27 & 6.50 
& 0.63 & 1.20 & 1.90 & 1.24  \\
& DriveArena~\cite{yang2024drivearena} & 8.89 & 17.52 & 0.29 & 9.78 & 10.93 & 7.00 & \underline{0.54} & \underline{1.04} & \underline{1.68} & \underline{1.09}  \\
& MagicDrive-V2~\cite{gao2025magicdrive} & \underline{12.68} & \underline{17.32} &  \underline{1.35} & \underline{14.37} & \underline{19.36} & \underline{11.69} & \underline{0.54} & 1.07 & 1.73 & 1.11  \\
& \textbf{\ourmodel~(Ours)} & \textbf{18.15} & \textbf{24.95} & \textbf{6.54} & \textbf{22.29} & \textbf{34.25} & \textbf{21.03} & \textbf{0.44} & \textbf{0.89} & \textbf{1.47} & \textbf{0.93} \\ \thickhline
\multirow{5}{*}{\rotatebox{90}{Rain}}
& Baseline & 31.60 & 41.20 & 25.50 & 25.47 & 23.28 & 24.75 & 0.35 & 0.70 & 1.18 & 0.75  \\
& Rule-based~\cite{li2019heavy} & 30.76 & 39.90 & 27.06 & 24.96 & 23.58 & 25.20 & 0.36 & 0.72 & 1.20 & 0.76 \\
& DriveArena~\cite{yang2024drivearena} & 33.46 & \textbf{45.25} & 29.19 & 32.66 & 25.32 & 29.06 & \underline{0.33} & \underline{0.65} & \underline{1.08} & \underline{0.71} \\
& MagicDrive-V2~\cite{gao2025magicdrive} & \underline{33.93} & 42.20 & \underline{31.96} & \underline{32.72} & \underline{25.39} & \underline{30.02} &0.34&	0.69&1.15&0.73 \\
& \textbf{\ourmodel~(Ours)} & \textbf{35.15} & \underline{44.26} & \textbf{33.13} & \textbf{33.00} & \textbf{28.04} & \textbf{31.39} & \textbf{0.29} & \textbf{0.55} & \textbf{0.90} & \textbf{0.58} \\ \thickhline

\end{tabular}}
\vspace{-15pt}
\label{tab:sparsedrive}
\end{table}

\begin{figure}[t]
\centering
\includegraphics[width=.99\linewidth]{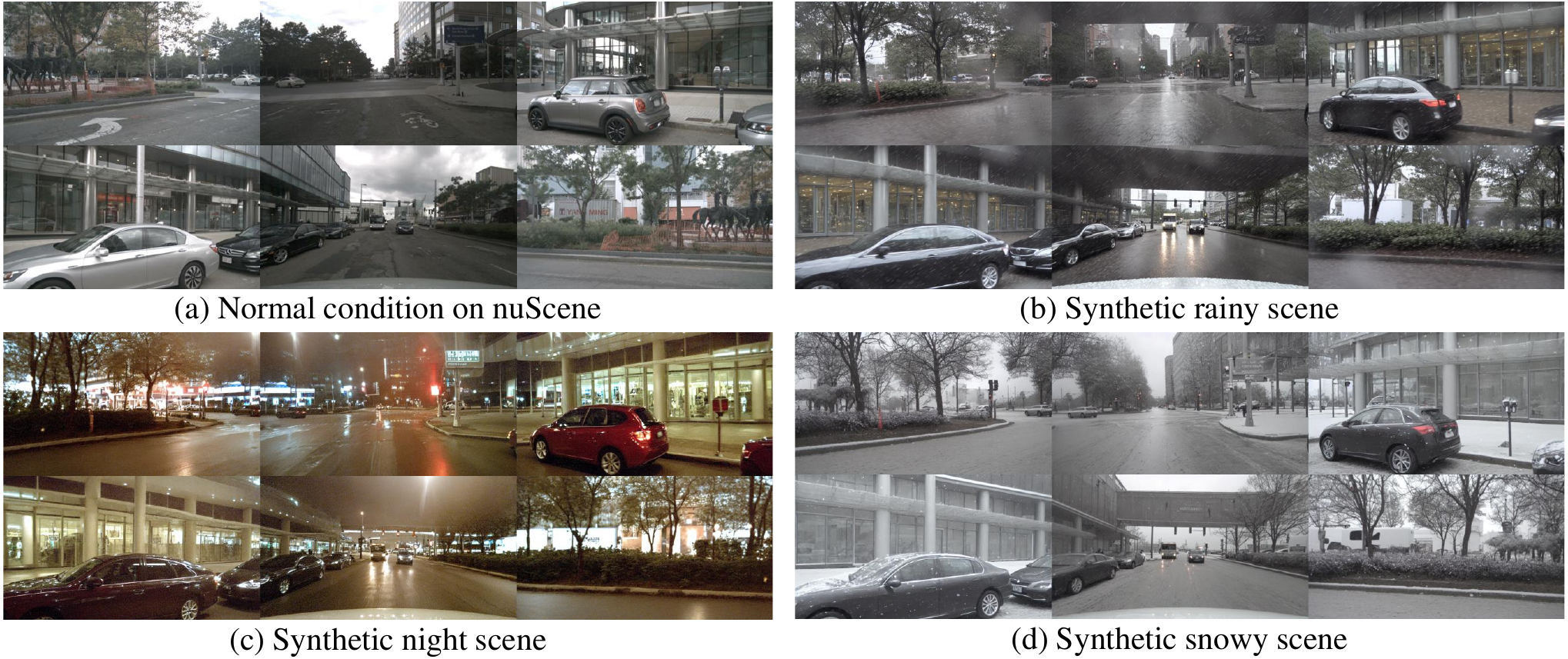}
\vspace{-5pt}
\caption{\textbf{Examples of text-driven weather-augmented samples.} Given text prompts specifying the target weather, conditioned on (a) normal scenes, \ourmodel~generates (b) synthetic rain, (c) synthetic night, and (d) synthetic snowy scenes. In the fifth view, raindrops on the car hood (rain) and strong light sources with their reflections on the road (night) are clearly generated.}
\label{fig:qual_weather}
\end{figure}

\noindent\textbf{Results.} In~\cref{tab:main}, our method achieves the strongest overall downstream performance among generative approaches, attaining the best 3D detection mAP/ NDS and BEV segmentation mIoU. The particularly strong BEV segmentation results indicate that our generated scenes exhibit high multi-view spatial consistency. 
We further assess generation fidelity with FVD: among SD-based image diffusion baselines, \ourmodel~achieves the lowest FVD, while STDiT-based video diffusion models still obtain lower FVD thanks to explicit temporal modeling.
At the same time, planning models trained on our generated data achieve the best metrics, closely matching the performance of models trained on real nuScenes data. Beyond superior quantitative results, our model also produces higher-quality fine-grained details. As shown in~\cref{fig:qual}, the competing method~\cite{yang2024drivearena} often removes scene elements or produces view-inconsistent object shapes, whereas our method better preserves the original scene and maintains strong multi-view consistency.
Additionally, we visualize our sequential generation results through~\cref{fig:qual_temporal}. Even in complex driving scenarios, our method generates highly competitive, temporally consistent images without relying on heavy video diffusion models.

\subsection{Data augmentation for autonomous driving}
\label{sec:adv}
We apply our domain-targeted augmentation with \ourmodel, driven solely by text prompts, to synthesize specific adverse conditions while preserving the original annotations. We evaluate on the nuScenes benchmark, focusing on the \textit{rain} and \textit{night} conditions. Based on the official metadata, we partition the dataset into three disjoint subsets: night, rain, and normal. For each target domain \(r \in \{\text{night}, \text{rain}\}\), we construct an augmented training set: given a normal sample \(x\), we synthesize its adverse-weather counterpart \(x^{r}\) with probability \(p\).

\noindent\textbf{Metric.}
We evaluate \ourmodel~against baselines by training SparseDrive~\cite{sun2025sparsedrive} on each augmented dataset. We report perception and planning performance under corresponding adverse conditions to quantify the improvements. 

\noindent\textbf{Baseline.}
We compare \ourmodel-based augmentation against three competitors: Baseline, which uses only original normal-weather samples, Rule-based methods~\cite{li2019heavy, thompson2002spatial}, which translate normal samples into night/rain images with hand-crafted filters, and DriveArena~\cite{yang2024drivearena}, which, like \ourmodel, 
a Stable Diffusion–based text-to-image synthesis framework, and MagicDrive-V2, a diffusion transformer (STDiT)–based model for text-conditioned data generation.


\noindent\textbf{Results.}
\Cref{tab:sparsedrive} reports the effect of data augmentation on perception and planning.
Under night conditions, the AD model trained with \ourmodel-augmented data attains the best overall performance, surpassing all baselines with higher detection and mapping scores and significantly lower planning errors.
A similar trend is observed under rainy conditions. These results indicate that the scene generation model trained with our knowledge-preserving scheme acquires stronger generative capability by text-prompting for rare adverse scenarios, which in turn leads to better autonomous driving performance under such conditions. 
Moreover, as illustrated in~\cref{fig:qual_weather}, our method can faithfully synthesize a broad range of weather conditions, including previously unseen scenarios such as snowy scenes.
Further details are provided in the \textit{supplementary materials.}


\section{Analysis}
\subsection{Ablation studies}
 
\noindent
\textbf{Knowledge-preserving spatio-temporal attention.}
We conduct toy ablations on our knowledge-preserving spatio-temporal attention (MISA and TRSA). As shown in~\cref{tab:ablation}, removing TRSA degrades temporal quality and worsens FVD while keeping BEV segmentation relatively high, whereas using only TRSA improves FVD but clearly harms BEV segmentation. The full model with both modules achieves the highest BEV mIoU and lowest FVD, confirming that MISA and TRSA are complementary for multi-view temporal continuity.

\noindent
\textbf{Object-presence ratio loss.}
We compare 3D object detection performance with and without the proposed object-presence ratio loss and break down the results by category. As shown in~\cref{tab:ablation_loss}, \ourloss~encourages more faithful object synthesis and yields consistent 3DOD gains, with particularly large improvements for rare categories such as motorcycle and bicycle (1.07\% and 1.00\% of the training set). This mitigates data imbalance in downstream detection. Additional per-category results and occurrence ratios are provided in the \textit{supplementary metarials.}




\subsection{Knowledge forgetting}
\label{sec:forgetting}
Prior studies have reported knowledge forgetting in personalization and fine-tuning of pretrained models~\cite{he2021analyzing, zhao2024continual, liao2025continual, kumari2023multi, ruiz2023dreambooth,ram2025dreamblend, mukhoti2023fine, aghajanyan2020better}. In particular, they indicate that fine-tuning can induce drift in the model’s text alignment~\cite{park2024textboost}.
Extending this observation to multi-view driving scene generation, we find that not only full/partial fine-tuning but also adding and training extra layers on top of a frozen backbone can induce similar drift.

\begin{figure}[t]
\centering
\includegraphics[width=.99\linewidth]{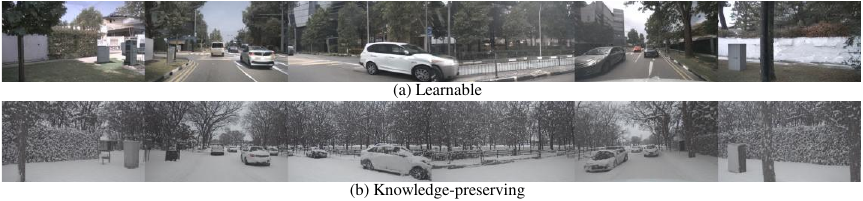}
\vspace{-7pt}
\caption{\textbf{Zero-shot text-guided generation results.}
Examples from a model with (a) learnable multi-view/temporal cross-attention and (b) our parameter-free knowledge-preserving spatio-temporal attentions with a ``Snowy weather" prompt.}
\vspace{-15pt}
\label{fig:qual_toy_ablation}
\end{figure}

\noindent
\textbf{Qualitative comparison.}
To isolate this effect, we construct a controlled baseline by replacing our parameter-free MISA and TRSA with learnable multi-view cross-attention and temporal cross-attention layers attached to the pretrained backbone (denoted as "SD + CA".)
\Cref{fig:qual_toy_ablation} (a) shows generation results with text prompt, "\textit{Snowy weather. Heavy snow.}"
While both our method and "SD + CA" successfully ensure multi-view and temporal consistency, "SD + CA" tends to degrade alignment and zero-shot controllability, indicating a forgetting of the pretrained prior. In contrast, our knowledge-preserving spatio-temporal attention enforces multi-view and temporal consistencies solely through input reshaping, while preserving text alignment.

\noindent
\textbf{Clip score.}
This behavior is further supported by the CLIP scores in~\cref{tab:clip_score}, which we employ as a relative measure to evaluate stylistic alignment against the reference samples. We compare our method against DriveArena~\cite{yang2024drivearena} and MagicDrive-V2~\cite{gao2025magicdrive}, while treating normal-weather nuScenes samples and real-weather data as references. For seen conditions such as night and rain, DriveArena \cite{yang2024drivearena} and MagicDrive-V2~\cite{gao2025magicdrive} achieve broadly comparable scores, suggesting that these models effectively mimic global weather textures. However, while these CLIP scores reflect stylistic "plausibility," our superior downstream performance in perception and planning (\cref{tab:sparsedrive}) provides more definitive evidence that our method not only preserves essential scene knowledge but also excels in the detailed realization of complex scene elements, such as precise road layouts and object instances. This distinction becomes even more pronounced in the snow setting, which is entirely absent from the training data. In this unseen scenario, our method surpasses existing generative baselines by a substantial margin, highlighting that our knowledge-preserving design is uniquely capable of generalizing to novel out-of-distribution conditions. 

\begin{table}[t]
\centering
\begin{minipage}[t]{0.49\linewidth}
\centering
\caption{\textbf{Ablation of knowledge-pre-serving spatio-temporal attention.} We evaluate BEV segmentation for multi-view consistency and FVD for temporal continuity.}
\vspace{-8pt}
\resizebox{0.99\linewidth}{!}{
\setlength{\tabcolsep}{4pt}
\begin{tabular}{cc|cccc|c}
    \hline
    \multicolumn{2}{c|}{Method} &
    \multicolumn{4}{c|}{BEV Segmentation mIoU (\%) ($\uparrow$)} &
    \multicolumn{1}{c}{Quality ($\downarrow$)} \\ \hline
    MISA & TRSA & Lanes & Drivable & Divider & Crossing & FVD \\ \hline
        $\checkmark$ & $\checkmark$ & \textbf{25.6} & \textbf{56.5} & \textbf{20.9} & \textbf{8.9} & \textbf{144.1} \\
    \hline
    $\checkmark$ & & 25.1 (\textcolor{red}{-0.5}) & 55.1 (\textcolor{red}{-1.4}) & 20.5 (\textcolor{red}{-0.4}) & 7.8 (\textcolor{red}{-1.1}) & 174.2 (\textcolor{red}{\textbf{+30.1}})\\
    & $\checkmark$ & 23.4 (\textcolor{red}{\textbf{-2.2}}) & 51.9 (\textcolor{red}{\textbf{-4.6}}) & 18.6 (\textcolor{red}{\textbf{-2.3}}) & 5.6 \textcolor{red}{(\textbf{-3.3}}) & 145.1 \textcolor{red}{(+1.0}) \\
    \hline
\end{tabular}}
\vspace{-7pt}
\label{tab:ablation}
\end{minipage}\hfill
\begin{minipage}[t]{0.49\linewidth}
\centering
\caption{\textbf{Ablation of object-presence ratio loss.} We evaluate 3DOD metric w/ and w/o the proposed loss, broken down by category. Categories are ordered left to right by decreasing frequency.}
\vspace{-5pt}
\resizebox{0.99\linewidth}{!}{
\setlength{\tabcolsep}{4pt}
\begin{tabular}{c|cc|cccc}
    \hline
    \multirow{2}{*}{Method} &
    \multicolumn{2}{c|}{3DOD ($\uparrow$)} &
    \multicolumn{4}{c}{Object category (AP, $\uparrow$)}  \\ \cline{2-7}
     & mAP & NDS & Car & Pedestrian & Motorcycle & Bicycle \\ \hline
     w/ loss  & \textbf{21.9} & \textbf{35.3} & \textbf{38.6} & \textbf{29.5} & \textbf{10.0} & \textbf{9.2} \\
     w/o loss & 16.1 (\textcolor{red}{-5.8}) & 29.6 (\textcolor{red}{-5.7}) & 33.5 (\textcolor{red}{-5.1}) & 25.3 (\textcolor{red}{-4.2}) & 0.4 (\textcolor{red}{-9.6}) & 1.9 (\textcolor{red}{-7.3}) \\
    \hline
\end{tabular}}
\vspace{-7pt}
\label{tab:ablation_loss}
\end{minipage}
\end{table}

\begin{table}[t]
\centering
    \begin{minipage}[t]{0.48\linewidth}
        \centering
        \caption{CLIP scores for weather-cond- itioned text–image generation.}
        \vspace{-5pt}
        \setlength{\tabcolsep}{12pt}
        \resizebox{0.95\linewidth}{!}{
        \begin{tabular}{l|c|c|c}
        \hline
        Method & Rain & Night & Snow \\ \hline
        Normal weather                                 &  0.2304 & 0.2319 & 0.1950 \\
        Real target weather                                    & 0.2564 & 0.2662 & 0.2711 \\ \hline
        DriveArena~\cite{yang2024drivearena}    & 0.2570 & 0.2716 & 0.2277 \\
        MagicDrive-V2~\cite{gao2025magicdrive}  & 0.2616 & 0.2705 & 0.2181 \\
        FrozenDrive (Ours)                      & 0.2595 & 0.2751 & 0.2507 \\
         \hline
        \end{tabular}}
        \label{tab:clip_score}
    \end{minipage}
    \hfill
    \begin{minipage}[t]{0.48\linewidth}
        \centering
        \caption{Lane alignment and Vbench~\cite{huang2024vbench} temporal consistency evaluation.}
        \vspace{-5pt}
        \setlength{\tabcolsep}{10pt}
        \resizebox{0.95\linewidth}{!}{
        \begin{tabular}{l|c|cc}
        \hline
        \multirow{2}{*}{Method} & Lane ($\uparrow$) & \multicolumn{2}{c}{Vbench ($\uparrow$)} \\ \cline{2-4}
        & mIoU & Subject & BG \\ \hline
        SD + CA & \underline{33.55} & 0.7586 & 0.8679 \\
        DriveArena~\cite{yang2024drivearena} & 30.73 & 0.7573 & 0.8556 \\
        MagicDrive-V2~\cite{gao2025magicdrive} & 26.71 & \textbf{0.7966} & \textbf{0.8845} \\
        FrozenDrive (Ours) & \textbf{33.58} & \underline{0.7724} & \underline{0.8700} \\ \hline
        \end{tabular}}
        \label{tab:consistency}
    \end{minipage}
\end{table}

\subsection{Consistency evaluation}
\label{sec:consistency}
While the previous ablation study provides indirect evidence of the consistency improvements brought by MISA and TRSA through BEV segmentation performance, we further investigate this aspect using two primary metrics that directly assess consistency. This complementary analysis allows us to examine the effect of MISA and TRSA on consistency more explicitly.

\noindent
\textbf{Lane Alignment.} 
To evaluate geometric consistency, we measure the degree of lane alignment, defined as the mIoU between the ground-truth and generated lanes in the image space. As shown in~\cref{tab:consistency}, our FrozenDrive provides geometric consistency comparable to the baseline that requires additional learnable cross-attention layers with partial fine-tuning (the same as "SD + CA" of~\cref{sec:forgetting}). This indicates that our method achieves high geometric accuracy without introducing additional parameters to the diffusion backbone.

\noindent
\textbf{Temporal Consistency.}
To assess temporal consistency, we adopt the subject and background consistency metrics from VBench~\cite{huang2024vbench}. Subject consistency is calculated based on DINO~\cite{caron2021emerging} feature similarities across frames to capture object-wise semantic correspondence, while background consistency utilizes CLIP~\cite{radford2021learning} similarities to evaluate global stability. 
Compared to MagicDrive-V2~\cite{gao2025magicdrive}, our method shows slightly lower consistency scores, which is expected since MagicDrive-V2 is built on a video diffusion model. Nevertheless, our FrozenDrive achieves stronger temporal consistency than the SD + CA and DriveArena \cite{yang2024drivearena}, both of which are fine-tuned baselines.

\subsection{Conclusion}
We propose \ourmodel, a zero-shot text-guided multi-view driving scene generator that steers a parameter-free frozen pretrained diffusion model. With our knowledge-preserving attention, \ourmodel~produces realistic, spatio-tempo-rally coherent multi-view driving scenes via driving-stack signals and text. Moreover, \ourmodel-augmented data improves downstream perception and planning, especially under rare and adverse conditions.

\noindent
\textbf{Limitations and future works.}
Our parameter-free frozen diffusion model offers good text alignment and controllability, but still lags behind recent video diffusion models in long-range temporal coherence. Moreover, existing protocols do not directly evaluate data augmentation quality, and developing more faithful metrics remains an important direction for future work. Extending our knowledge-preserving design beyond Stable Diffusion to stronger video diffusion or DiT-based backbones is another promising direction.

 




\newcommand{\xmark}{\ding{55}}

\renewcommand{\thetable}{\Alph{table}}
\renewcommand{\thefigure}{\Alph{figure}}
\renewcommand{\theequation}{\Alph{equation}}
\renewcommand{\thesection}{\Alph{section}}
\renewcommand{\thesubsection}{\thesection.\Alph{subsection}}

\makeatletter
\def\thickhline{%
  \noalign{\ifnum0=`}\fi\hrule \@height \thickarrayrulewidth \futurelet
   \reserved@a\@xthickhline}
\def\@xthickhline{\ifx\reserved@a\thickhline
               \vskip\doublerulesep
               \vskip-\thickarrayrulewidth
             \fi
      \ifnum0=`{\fi}}
\makeatother
\setlength{\thickarrayrulewidth}{2.5\arrayrulewidth}


\title{FrozenDrive: Zero-Shot Text-Guided Driving Scene Generation and Data Augmentation with Parameter-Free Frozen Diffusion Model \\ \textmd{-Supplementary material-}}

\titlerunning{Abbreviated paper title}

\author{Yuhwan Jeong\orcidlink{0009-0002-0279-146X}\thanks{Equal contribution.} \and
Hyeonseong Kim\orcidlink{0009-0003-9792-4647}\protect\CoAuthorMark \and
Daehyun We\orcidlink{0009-0003-5652-1681}\protect\CoAuthorMark \and
Seonkyu Song\orcidlink{0009-0003-8282-669X}\protect\CoAuthorMark \and
Jinnyeong Yang\orcidlink{0009-0002-9275-6296}\protect\CoAuthorMark \and
Hyun-Kurl Jang\orcidlink{0009-0003-7943-3326} \and
Youngho Yoon\orcidlink{0009-0003-4346-8260} \and
Kuk-Jin Yoon\orcidlink{0000-0002-1634-2756}}

\authorrunning{F.~Author et al.}


\institute{KAIST, Visual Intelligence Lab \\
\email{\{jeongyh98, brian617, greathyun, sok9854, jinnyeong6118, jhg0001, dudgh1732, kjyoon\}@kaist.ac.kr}}




\maketitle

\section{Overview}
\label{sec:rationale}
\noindent
Our supplementary material offers further insights into the proposed method and provides in-depth discussions on topics that were not extensively covered in the main paper:

\begin{itemize}
    \item Object imbalance (Sec.~\ref{sec:object_loss_sup}).
    \item Implementation details (Sec.~\ref{sec:imple_supple}).
    \item Data augmentation (Sec.~\ref{sec:da_supple}).
    \item Knowledge forgetting (Sec.~\ref{sec:forgetting_supple}).
    \item Additional qualitative results (Sec.~\ref{sec:more_qual}).
\end{itemize}


    

%

\section{Object imbalance}
\label{sec:object_loss_sup}
\noindent\textbf{Object occurrence ratio.}
In a supervised setting, the fidelity of generated images largely depends on how frequently samples are observed during optimization. 
\Cref{tab:object_amount} reports per-class instance counts for the nuScenes training set~\cite{caesar2020nuscenes}, revealing a pronounced imbalance: \emph{cars} dominate, whereas \emph{bicycles} appear at only 0.023 times the car frequency, making them considerably harder to learn and render faithfully.

\begin{table}[t]
\centering
\caption{\textbf{Category-wise object counts and ratios in the nuScenes training dataset.} The `Ratio vs. car’ column denotes, for each object category, its ratio relative to the most frequent category, car.}
\resizebox{0.7\linewidth}{!}{
\setlength{\tabcolsep}{6.0pt}
\begin{tabular}{l|rrr}
\hline
\multicolumn{1}{l|}{Object} & \multicolumn{1}{c}{Amounts} & Ratio (\%) & Ratio vs. `car' (\%)  \\ \hline
car                  & 413,318 & 43.74 & 100 \\
pedestrian           & 185,847 & 19.67 & 44.96\\
barrier              & 125,095 & 13.23 & 30.27\\
traffic cone         & 82,362  & 8.72 & 19.93 \\
truck                & 72,815  & 7.71 & 17.62\\
trailer              & 20,701  & 2.19 & 5.00 \\
bus                  & 13,163  & 1.39 & 3.18\\
construction vehicle & 11,993  & 1.27 & 2.90\\
motocycle            & 10,109  & 1.07 & 2.45\\
bicycle              & 9,478   & 1.00 & 2.29\\ \hline
total                & 944,881 & 100.00 & - \\ \hline
\end{tabular}}
\label{tab:object_amount}
\end{table}

\noindent\textbf{More details of object presence ratio loss.} Based on this statistical analysis, we propose the object-presence ratio, as defined in~Eq.~{\color{red}6} and~{\color{red}7} of the main paper. We calculate class-dependent weight as follows:

\begin{equation}
    w_k =
    \begin{cases}
    ({{o_t}/{o_k}})_p, & \text{if }  \mathcal{K}_p \neq \emptyset,\\
    0,                                & \text{if } \mathcal{K}_p = \emptyset,
    \end{cases}
\end{equation}
where $o_t$ and $o_k$ denote the amounts of total objects and class $k$ objects, respectively. 
For example, when a pixel $p$ belongs to both the \emph{car} and \emph{bus} labels, the corresponding class weights are $w_{\text{car}} = 2.29$ and $w_{\text{bus}} = 71.78$. 
According to the max rule, the pixel weight is given by
\begin{equation}
w(p) = \max\{w_{\text{car}}, w_{\text{bus}}\} = 71.78.
\end{equation}
In contrast, for another pixel $q$ that is not assigned to any class 
(i.e., $\mathcal{K}_q = \emptyset$), we set its weight to $w(q) = 0.$ For the hyperparameter $\lambda$, we set it to 0.02.

\noindent\textbf{Performance breakdown by category.}
To address object category imbalance, we enforce an object-focused constraint on rare categories, guiding the model to devote more capacity to underrepresented objects and enhancing their generative quality instead of overemphasizing frequent classes.
\Cref{tab:uniad_ablation} shows the per-class 3D detection AP of the UniAD-based detector trained on nuScenes~\cite{caesar2020nuscenes}, comparing DriveArena~\cite{yang2024drivearena}, MagicDrive-V2~\cite{gao2025magicdrive}, our \ourmodel\ without the proposed \ourloss, and the full \ourmodel.
With \ourloss, \ourmodel\ improves not only frequent, large classes (\eg \emph{car}, \emph{truck}) but also rarer or smaller ones (\eg \emph{bus}, \emph{trailer}, \emph{pedestrian}, \emph{bicycle}, \emph{motorcycle}).
Even without this constraint, our model already surpasses the baselines in terms of mAP, and adding \ourloss\ yields a substantial additional gain, significantly outperforming all compared methods in most categories.
These gains indicate that the proposed \ourloss\ effectively rebalances learning toward scarce categories, which in turn translates into stronger downstream detection performance.

\section{Implementation details}
\label{sec:imple_supple}
\subsection{Training and inference}
%
%
%
%
%
%
We initialize the diffusion backbone with Stable Diffusion v1.5~\cite{rombach2022high} and keep all backbone weights frozen with no additional layers or trainable parameters added. The ControlNet~\cite{zhang2023adding} and the condition embedders (Sec.~\ref{sec:embedding}) are randomly initialized and trained from scratch. Training uses AdamW~\cite{LoshchilovH19} optimizer with a learning rate of $1{\times}10^{-4}$ for $200$K iterations and batch size $4$, following a two-stage resolution schedule: $150$K iterations at $224{\times}400$ followed by $50$K at $448{\times}800$, which provides efficient convergence at high resolution. All training is performed on two NVIDIA A100 GPUs. At training time, we enable multi-view inflated self-attention to learn cross-view coherence with the frozen backbone; at inference, we additionally activate temporal reference self-attention, using the most recently generated frame ($j{=}1$) as reference, to produce spatio-temporally consistent multi-view sequences. 
The final frames are synthesized at $448{\times}800$ and upsampled by bilinear interpolation to the original nuScenes resolution ($900{\times}1600$) for downstream agents such as UniAD~\cite{hu2023planning} and SparseDrive~\cite{sun2025sparsedrive}.
The generation resolutions of the compared methods in the main experiment (Tab.~{\color{red}1} of the main paper) are summarized in~\cref{tab:resolution}.

Our full model comprises 1.40B parameters, of which only 0.37B are trainable, as we freeze the backbone and train only the embedding and ControlNet modules.
During inference, FrozenDrive operates at a resolution of 1 × 448 × 800 and requires only a single A100 GPU, using 12.43GB of memory and taking 0.97s per frame per step in the 1-GPU setting. In comparison, MagicDrive-V2, evaluated using its official codebase, operates at a resolution of 6 × 848 × 1600 and requires 4 A100 GPUs, using 59.95GB of memory per GPU and taking 0.77s per frame per step in the 4-GPU setting.

\begin{table}[t!]
\centering
\caption{\textbf{3D object detection performance with per-class AP (\%) and mean AP.} The proposed \ourloss~improves not only frequent, large classes (\eg car and truck) but also rarer or smaller ones (\eg bus, trailer, motorcycle, and bicycle).}
\label{tab:uniad_ablation}
\setlength{\tabcolsep}{2.0pt}
\resizebox{\linewidth}{!}{
\begin{tabular}{l*{11}{c}}
\toprule
\multirow{2}{*}{Method} & \multicolumn{10}{c}{Class AP (\%)} & \multirow{2}{*}{\makecell{mAP (\%)}} \\
\cmidrule(lr){2-11}
 & car & truck & bus & trailer & construction & pedestrian & motorcycle & bicycle & traffic\,cone & barrier & \\
\midrule
nuScenes~\cite{caesar2020nuscenes}       & 56.89 & 32.00 & 38.50 & 14.87 & 11.41 & 43.95 & 38.76 & 36.07 & 55.27 & 52.22 & 37.99 \\
\midrule
DriveArena~\cite{yang2024drivearena} & 29.82 &  7.75 &  9.41 &  3.33 &  0.01 & 14.74 &  3.58 &  0.42 & 37.70 & 41.75 & 14.85 \\
MagicDrive-V2~\cite{gao2025magicdrive} & 38.00 & \textbf{9.40} & 7.00 & 3.20 & 0.00 & 17.60 & 3.00 & 0.20 & 33.20 & 40.80 & 15.24 \\
\ourmodel~(w/o loss) & 33.50 & 5.50 & 1.20 & 2.70 & 0.00 & 25.30 & 0.40 & 1.90 & 49.10 & 41.90 & 16.15 \\
\ourmodel~(Ours)       & \textbf{38.60} &  8.60 & \textbf{17.50} &  \textbf{6.20} &  \textbf{0.70} & \textbf{29.50} & \textbf{10.00} &  \textbf{9.20} & \textbf{49.70} & \textbf{48.70} & \textbf{21.87} \\
\bottomrule
\end{tabular}}
\vspace{-10pt}
\end{table}


\subsection{Conditional embedding}
\label{sec:embedding}
%
We guide the diffusion denoising process toward high-fidelity driving scenes using five conditioning signals: (i) a scene layout (derived from an HD map and 3D bounding boxes), (ii) a depth map, (iii) a per-view camera indicator, (iv) the relative pose to the previous frame, and (v) a text description.
Each non-text signal $c_k$ is mapped by a lightweight embedding network $E_k$ to a latent-space embedding $\mathbf{e}_k = E_k(c_k)$, which is combined with the latent feature $\mathbf{z}$ and passed to ControlNet:
\begin{equation}
\mathrm{ControlNet}\!\left(\mathbf{z}+\sum_{k\in\{\text{layout,\,depth,\,view,\,pose}\}}{\mathbf{e}_k}\right).
\end{equation}
Text description is encoded by the frozen CLIP text encoder~\cite{radford2021learning} and used in cross-attention. Below, we detail the architectures of the embedding networks $E_k$ and the condition-processing pipeline for each signal.

\noindent\textbf{Scene layout.} 
%
%
We construct a per-view scene-layout map by projecting HD map elements and 3D object bounding boxes onto each image plane, as in DriveArena~\cite{yang2024drivearena}. Projection uses the camera intrinsics and extrinsics of each view. We treat every HD map layer (\emph{divider}, \emph{pedestrian crossing}, \emph{boundary}, \emph{drivable}) as a separate category, alongside all 3D bounding box categories (see object categories of \cref{tab:object_amount}). Each category is rasterized into a binary mask, and stacking all category masks channel-wise forms the layout tensor for the view. The resulting multi-channel map is encoded by the conditional encoding network~\cite{zhang2023adding} (\ie $E_{\text{layout}}$).

\begin{table}[t]
\centering

\begin{minipage}[t]{0.4\linewidth}
\centering
\captionof{table}{View index for view embedding. Each view index indicates the corresponding camera view.}
\resizebox{\linewidth}{!}{
\setlength{\tabcolsep}{8.0pt}
\begin{tabular}{c|ccc}
\hline
\multirow{2}{*}{View} & Front & Front & Front \\
& left & center & right \\
\hline
Index & 0 & 1 & 2 \\
\hline
\multirow{2}{*}{View} & Back & Back & Back \\
& right & center & left \\
\hline
Index & 3 & 4 & 5 \\
\hline
\end{tabular}}
\label{tab:view_index}
\end{minipage}
\hfill
\begin{minipage}[t]{0.58\linewidth}
\centering
\captionof{table}{Scene-wise split of the nuScenes validation set under adverse-weather conditions. We sample 12 scenes and use them as evaluation scenes.}
\resizebox{\linewidth}{!}{
\setlength{\tabcolsep}{6.0pt}
\begin{tabular}{crrrrrr}
\hline
Scene type & \multicolumn{6}{c}{Scene number} \\
\hline
\multirow{2}{*}{Night}
  & 1059 & 1061 & 1062 & 1063 & 1064 & 1066 \\
  & 1068 & 1069 & 1070 & 1071 & 1072 & 1073 \\
\hline
\multirow{2}{*}{Rain}
  & 0626 & 0632 & 0634 & 0635 & 0636 & 0637 \\
  & 0638 & 0905 & 0907 & 0908 & 0911 & 0912 \\
\hline
\end{tabular}}
\label{tab:scene_split}
\end{minipage}

\end{table}

\begin{table}[t]
\centering
\caption{Distribution of train scenes.}
\setlength{\tabcolsep}{10pt}
\renewcommand{\arraystretch}{1.3}

\resizebox{0.45\linewidth}{!}{
\begin{tabular}{c|c c c}
\hline
Scene type & Normal & Night & Rain \\
\hline
\# Sample & 19685 & 2863 & 5060 \\
\hline
\end{tabular}}
\label{tab:prob}
\end{table}

\noindent\textbf{Depth map.}
%
To provide explicit 3D geometric guidance, we use a depth map as a condition. We first temporally aggregate LiDAR point clouds into a binary 3D occupancy grid~\cite{li2025uniscene}. For each camera view, depth is then computed on the image plane by ray casting with the corresponding camera intrinsics and extrinsics parameters; a $z$-buffer selects the nearest occupied voxel along each ray. The resulting per-view depth map is clipped to a fixed range ($50$m) and min-max normalized to $[0,1]$. Invalid pixels (free space) are filled with $-1$. Finally, the normalized depth map is encoded by the conditional encoding network ($E_{\text{depth}}$) to produce the depth embedding.

\noindent\textbf{Camera view.}
%
To indicate which camera view produced each image, we encode a per-view identifier as a conditioning signal. Each camera view is assigned a unique index $v \in \{0,\dots,n_{\text{view}}-1\}$ (see \cref{tab:view_index}), which we embed using Gaussian Fourier features~\cite{tancik2020fourier} (\ie $E_{\text{view}}$). 
The resulting feature is broadcast spatially before being passed to ControlNet.

\noindent\textbf{Relative pose.}
%
We encode the relative pose between the current frame and the reference frame as a 2D spatial embedding. For each pixel coordinate $(x,y)$ on the image plane, we first lift it to a 3D point $(x,y,0)$ and apply the $4{\times}4$ relative pose transform to obtain a warped 3D coordinate. We then take the in-plane components $(x',y')$ and map them with Gaussian Fourier features~\cite{tancik2020fourier} (\ie $E_{\text{pose}}$) to produce a per-pixel feature. The resulting feature map is passed through a linear projector and is added to the latent feature before being fed to ControlNet.



\begin{figure}[t]
    \centering
    \captionsetup{type=figure}
    \includegraphics[width=.99\linewidth]{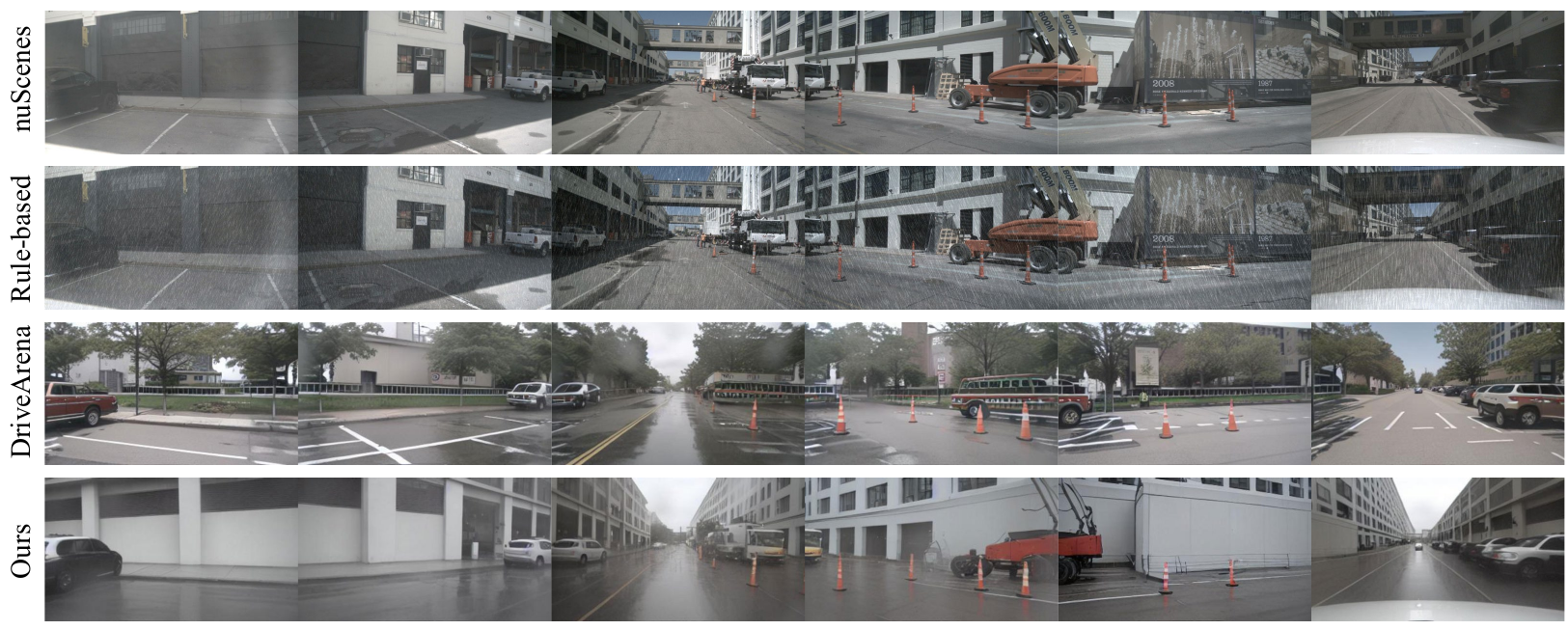}
    \vspace{-5pt}
    \caption{The rain-condition samples used in the DA dataset.}
    \vspace{-5pt}
    \label{fig:DA_rain}
\end{figure}

\begin{figure}[t]
    \centering
    \captionsetup{type=figure}
    \includegraphics[width=.99\linewidth]{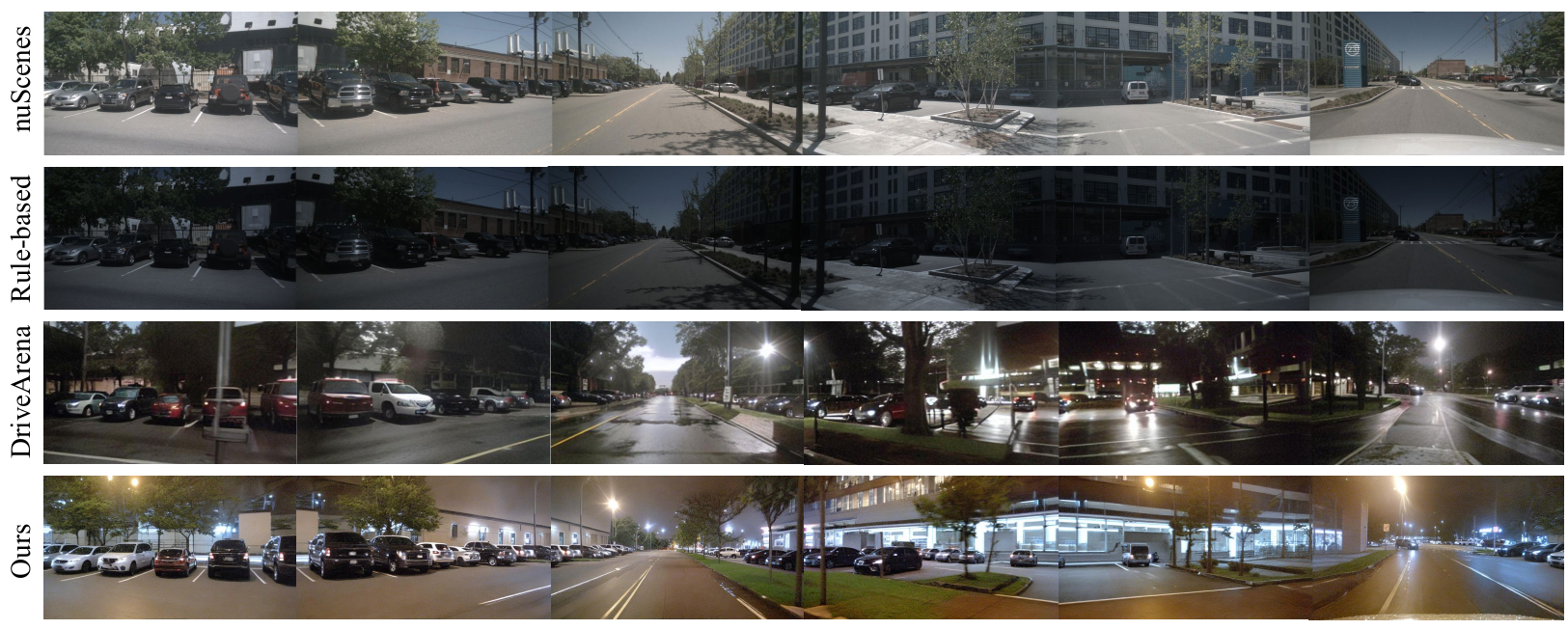}
    \vspace{-5pt}
    \caption{The night-condition samples used in the DA dataset.}
    \vspace{-15pt}
    \label{fig:DA_night}
\end{figure}

\section{Data augmentation}
\label{sec:da_supple}

\subsection{Data augmentation for autonomous driving}
\noindent\textbf{Prepare dataset.}
%
To validate the effectiveness of our data augmentation for end-to-end AD models, we constructed a custom evaluation set from the official nuScenes validation split, consisting only of scenes corresponding to the target adverse conditions.
The validation set was partitioned on a per-scene basis, where each scene was categorized into a specific adverse condition (\eg, night or rain) according to the scene-level descriptions provided by the dataset.
For instance, the scene “scene-1061” is labeled with the description “Night, turn, trash can, residential” and was therefore assigned to the night-condition subset. 
To clearly separate night and rain domains, we excluded any overlapping scenes that belong to the intersection of the night and rain conditions from the nuScenes validation set.
For each adverse-weather domain, we sampled 12 scenes and used them as evaluation scenes. The indices of the selected scenes are summarized in \cref{tab:scene_split}. 

\noindent\textbf{Training procedure.}
During training, at each iteration, we randomly replace a normal scene with its adverse-weather counterpart $x^{r}$ with probability $p_r$, and train separate AD models for each target domain $r \in \{\text{night}, \text{rain}\}$. The sampling probability $p_r$ for each target domain is computed as:
\begin{equation}
p_{r} = {\text{sample}_{r}}/{\text{sample}_{\text{normal}}\vphantom{gp}},
\label{eq:pr}
\end{equation}
where the resulting values for night and rain conditions are $0.15$ and $0.26$, respectively.
Sample scenes from the training set are illustrated in~\cref{fig:DA_rain} for rain and in~\cref{fig:DA_night} for night.

For training SparseDrive, we adopt a two-stage pipeline that first learns the perception module under normal conditions and then optimizes the planning module on top of the learned representations. In the pre-stage, we train the perception network for 100 epochs using only samples from the normal split, and regard the resulting checkpoint as our baseline model. This baseline serves as a reference point for all subsequent experiments and allows us to isolate the effect of introducing adverse-weather data during later stages of training.

Building on this baseline, we perform stage~1 training by fine-tuning the perception module for an additional 20 epochs using an augmented dataset that includes adverse-weather samples (e.g., night and rain). This step exposes the model to a broader range of visual conditions while retaining the structure learned from normal scenes. In stage~2 (planning training), we attach and train the planning module for both the synthesized model and other compared methods, each initialized from its corresponding stage-1 checkpoint. Both variants are trained using the same augmented dataset, ensuring a fair comparison of planning performance under identical training conditions. The training settings for each stage are summarized in~\cref{tab:training_hyperparams}.

\noindent\textbf{Result analysis.} 
Table~{\color{red}2} of the main paper reports the effect of data augmentation on perception and planning. Under night conditions, the AD model trained with \ourmodel-augmented data attains the best overall performance, surpassing all baselines with substantially higher 3D detection and online mapping scores and clearly lower planning errors. A closer look at the night results shows particularly large gains in both mAP and NDS, suggesting that the night scenes synthesized by \ourmodel~more closely resemble real nighttime imagery than those produced by rule-based filters, DriveArena, or MagicDrive-V2. By preserving the original scene geometry while realistically rendering low-light appearance, our synthetic data exposes the model to richer object appearances and road structures, which in turn facilitates learning robust 3D detectors and more accurate map predictions. This improvement in perception quality directly translates into smaller trajectory errors and lower collision rates for the planner.

A similar but milder trend is observed under rainy conditions, where \ourmodel~still consistently improves detection, mapping, and planning metrics over all baselines. These results indicate that the scene generation model trained with our knowledge-preserving scheme acquires stronger generative capability for rare adverse scenarios via text prompting, thereby enhancing downstream autonomous driving performance.

\subsection{Prompt-based augmentation}

\noindent\textbf{Adverse weather CLIP score.}
To assess whether the synthesized scenes faithfully capture the intended style specified primarily by the weather text prompts, we evaluate the generated scenes using a CLIP-based metric. For a given input text, we first generate a scene and then construct a set of 100 text descriptions associated with that text, and compute a correlation score between the scene and these texts based on their CLIP similarities. 
The 100 reference sentences are generated using GPT, conditioned only on common factors related to weather attributes and on-road driving scenarios. 
Since there are not enough multi-view sequences for each weather condition, we extract only front-view images and compute CLIP scores on this view.
For the real-weather references, night-time images are taken from the Waymo~\cite{sun2020scalability} and DSEC~\cite{gehrig2021dsec} datasets, rainy scenes from Waymo, and snowy scenes from CADC~\cite{pitropov2021canadian}.

\begin{figure}[t]
    \centering
    \captionsetup{type=figure}
    \includegraphics[width=.99\linewidth]{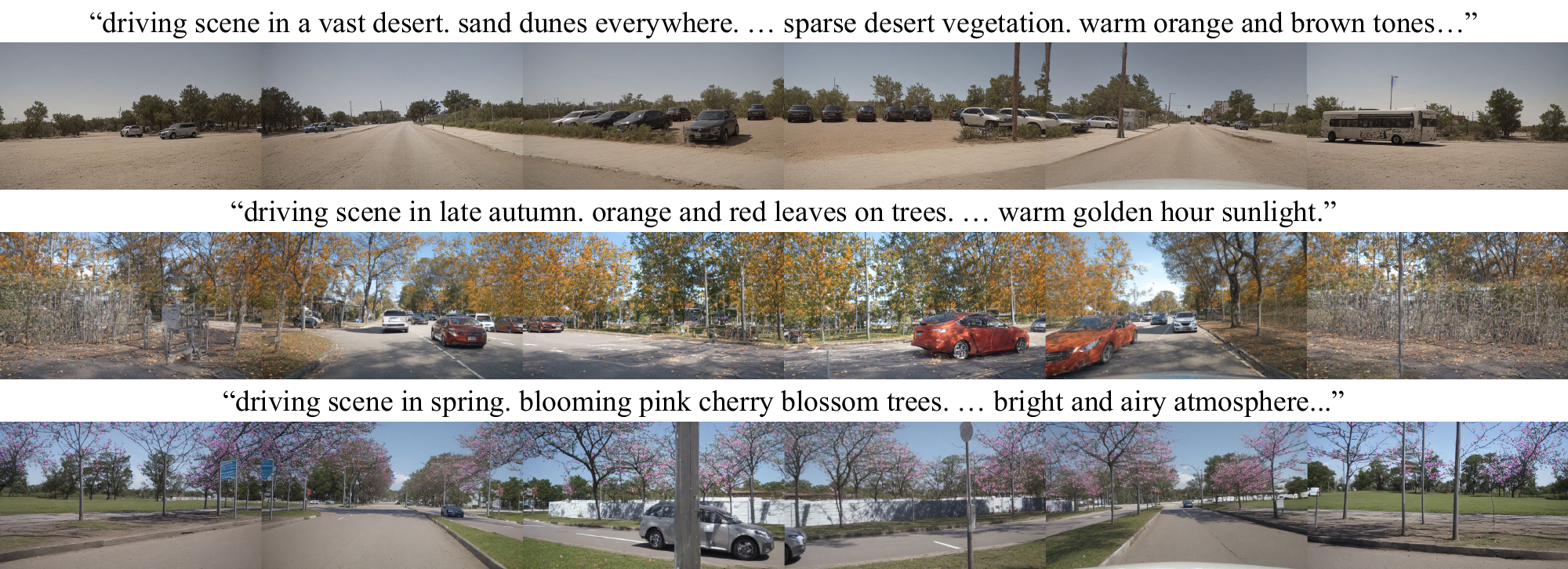}
    \vspace{-10pt}
    \caption{Examples of generated images from various text prompts.}
    \vspace{-10pt}
    \label{fig:prompt}
\end{figure}

\begin{figure*}
    \centering
    \captionsetup{type=figure}
    \includegraphics[width=.99\linewidth]{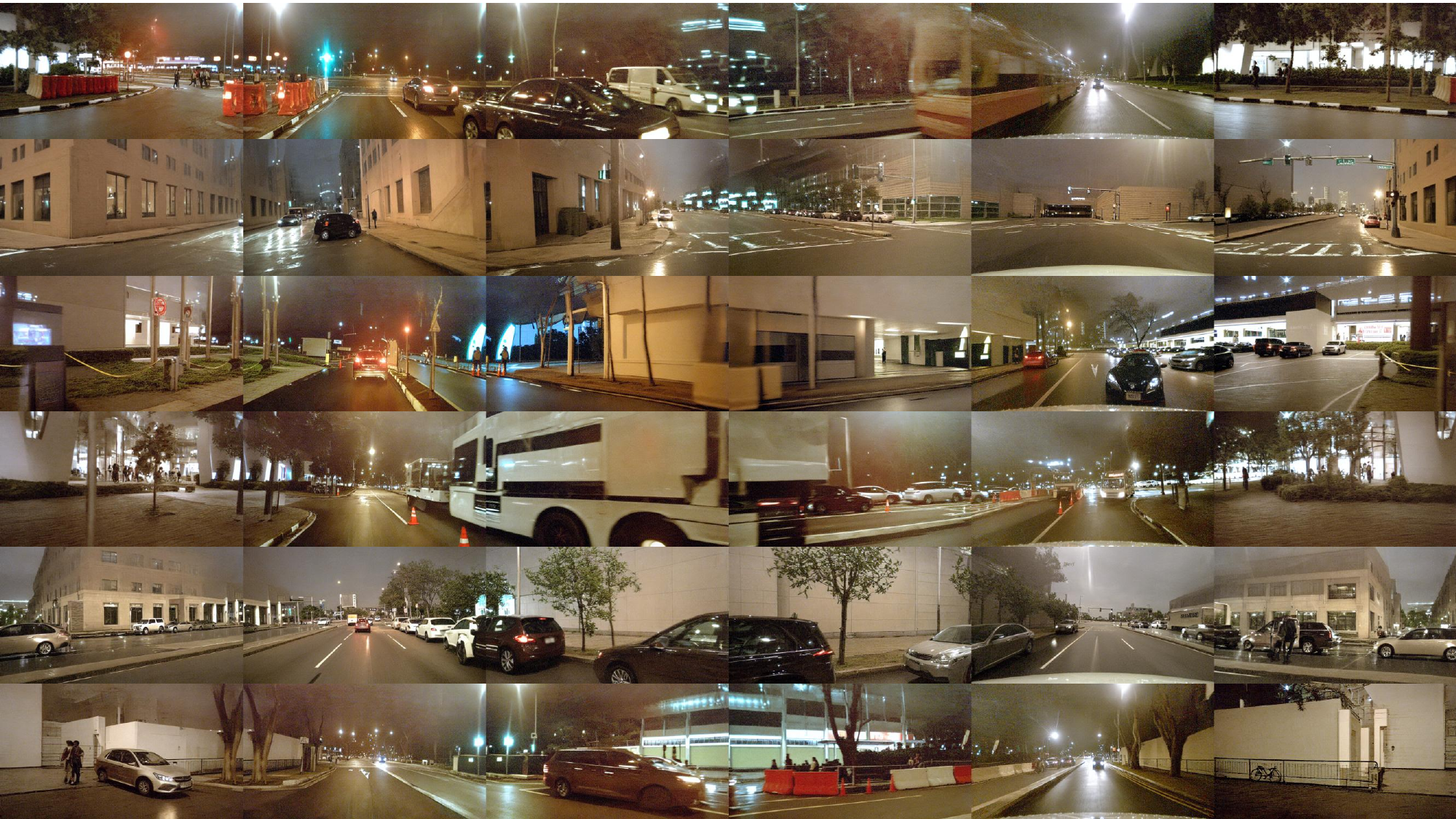}
    \vspace{-5pt}
    \caption{Example scenes of night condition augmentations.}
    \label{fig:more_weather_night}
\end{figure*}

\begin{figure*}
    \centering
    \captionsetup{type=figure}
    \includegraphics[width=.99\linewidth]{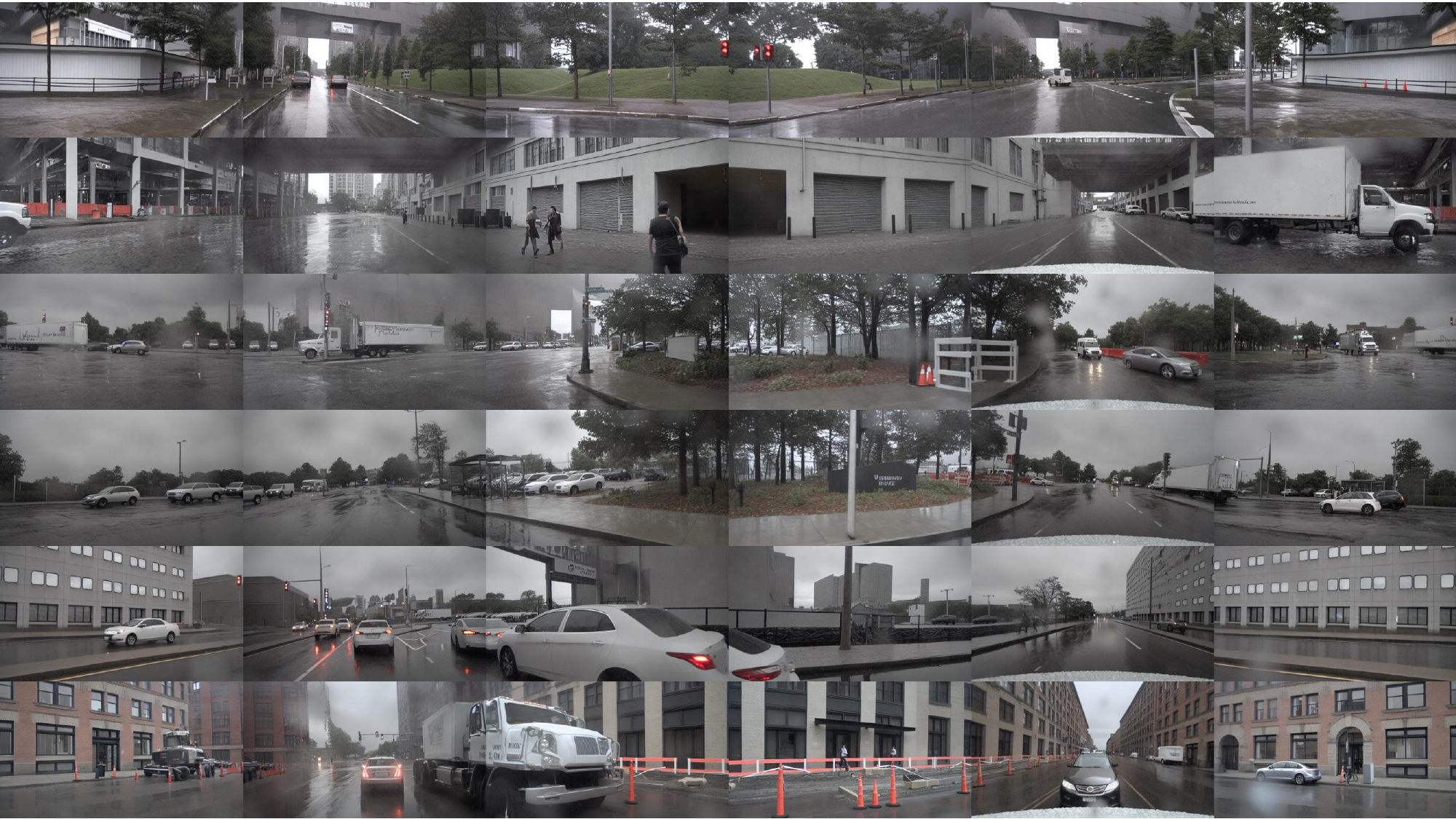}
    \vspace{-5pt}
    \caption{Example scenes of rainy weather augmentations.}
    \label{fig:more_weather_rain}
\end{figure*}

\begin{figure*}
    \centering
    \captionsetup{type=figure}
    \includegraphics[width=.99\linewidth]{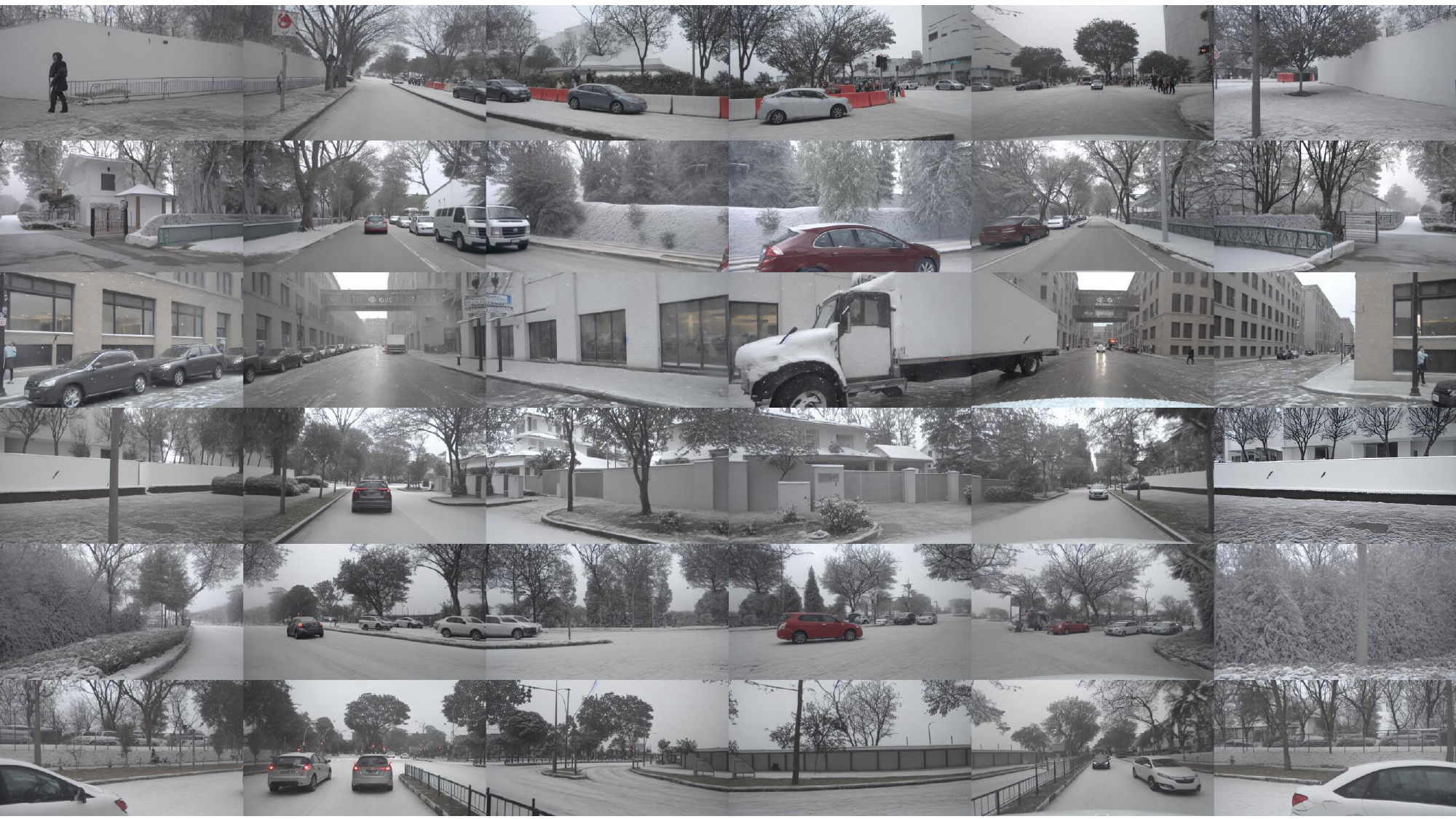}
    \vspace{-5pt}
    \caption{Example scenes of snowy weather augmentations.}
    \label{fig:more_weather_snow}
\end{figure*}

\noindent\textbf{Text prompting.}
We provide several visualizations of diverse scenes generated from various text prompts. 
\Cref{fig:prompt} provides examples of our scene synthesis results, highlighting how different prompts induce consistent and semantically plausible changes in weather and style while preserving the overall scene structure.
\Cref{fig:more_weather_night}, \ref{fig:more_weather_rain}, and \ref{fig:more_weather_snow} present extra visualizations for our main weather targets (night, rain, and snow), which are also used in our CLIP-score evaluation.

\begin{figure}[t]
    \centering
    \captionsetup{type=figure}
    \includegraphics[width=.98\linewidth]{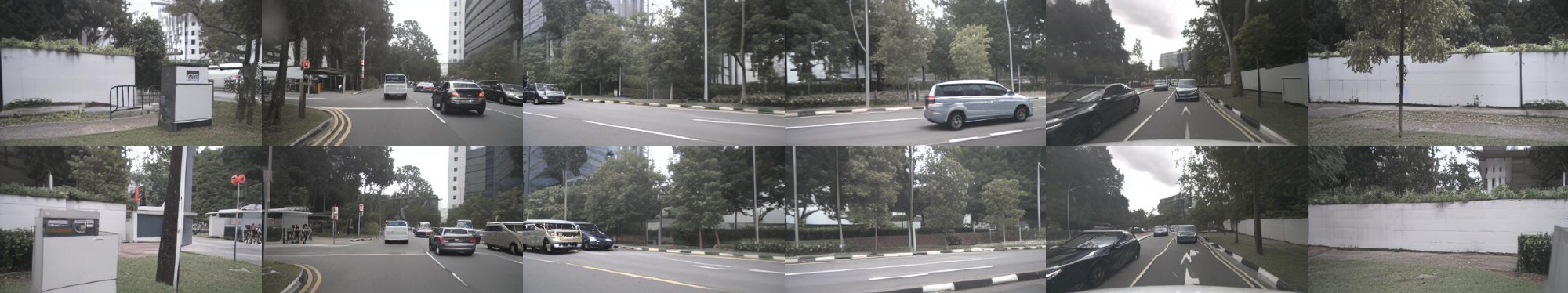}
    \vspace{-5pt}
    \caption{\textbf{Zero-shot text-guided generation results with learnable multi-view cross-attention.} Adding learnable multi-view cross-attention improves cross-view consistency but degrades text alignment. Input prompt: ``Snowy weather. Heavy snow.''}
    \label{fig:spatial_forgetting}
\end{figure}


\begin{table}[t]
\centering
\caption{Training settings of SparseDrive for adverse-weather-conditioned driving.}
\setlength{\tabcolsep}{9pt}
\resizebox{0.6\linewidth}{!}{
\begin{tabular}{c|c|c|c}
\hline
  & Batch Size & Epochs & Lr \\ \hline
Pre-stage & 32 & 100 & $1 \times 10^{-4}$ \\
Stage 1 & 32 & 20 & $1 \times 10^{-4}$  \\
Stage 2 & 24 & 10  & $7.5 \times 10^{-5}$  \\ \hline
\end{tabular}}
\label{tab:training_hyperparams}
\end{table}

\begin{table}[t]
\centering
\caption{Resolution used by each method ($T \times H \times W$).}
\setlength{\tabcolsep}{9pt}
\resizebox{0.5\linewidth}{!}{
\begin{tabular}{c|c}
\hline
Method & Resolution  \\ \hline
MagicDrive \cite{gao2024magicdrive} & 1 x 224 × 400 \\ 
Panacea\cite{wen2024panacea} & 8 x 256 × 512 \\ 
DrivingSphere\cite{yan2025drivingsphere} & 16 x 1080 x 1920 \\ 
DiST-4D\cite{guo2025dist}  & 17 x 424 x 800 \\
DriveArena\cite{yang2024drivearena} & 1 x 224 × 400 \\
MagivDrive-V2\cite{yang2024drivearena} & 6 x 848 × 1600 \\
X-Scene\cite{yang2025xscene} & 7 x 224 × 400 \\
FrozenDrive (Ours)  & 1 x 448 x 800 \\ \hline
\end{tabular}}
\label{tab:resolution}
\end{table}

\section{Knowledge forgetting analysis details}
\label{sec:forgetting_supple} 
In the main paper (Sec.~\textcolor{red}{5.2}), we showed that replacing our parameter-free multi-view inflated self-attention (MISA) and temporal reference self-attention (TRSA) with \emph{learnable} cross-attention layers induces forgetting of the pretrained diffusion prior. Concretely, for multi-view coherence, we attached multi-view cross-attention that queries each view over neighboring-view latents (as in Drive-Arena~\cite{yang2024drivearena}), and for temporal consistency, we attached temporal cross-attention that queries the current frame over a reference frame. We further observe forgetting even when only one of these modules is added. As illustrated in \cref{fig:spatial_forgetting}, adding learnable multi-view cross-attention improves cross-view agreement but \emph{degrades text alignment} and offers limited temporal coherence: under the prompt “Snowy weather. Heavy snow,” the model under-reflects the text, yielding weak snow cues. These results support our claim that training extra attention layers atop a pretrained backbone erodes the text–image prior, whereas our knowledge-preserving spatio-temporal attention enforces cross-view and temporal consistency by reshaping inputs without new trainable backbone layers, thereby maintaining strong text alignment (main paper Fig.~\textcolor{red}{7}).

\section{More visualizations}
\label{sec:more_qual}
\begin{figure*}[p]
    \centering
    \captionsetup{type=figure}
    \includegraphics[width=.99\linewidth]{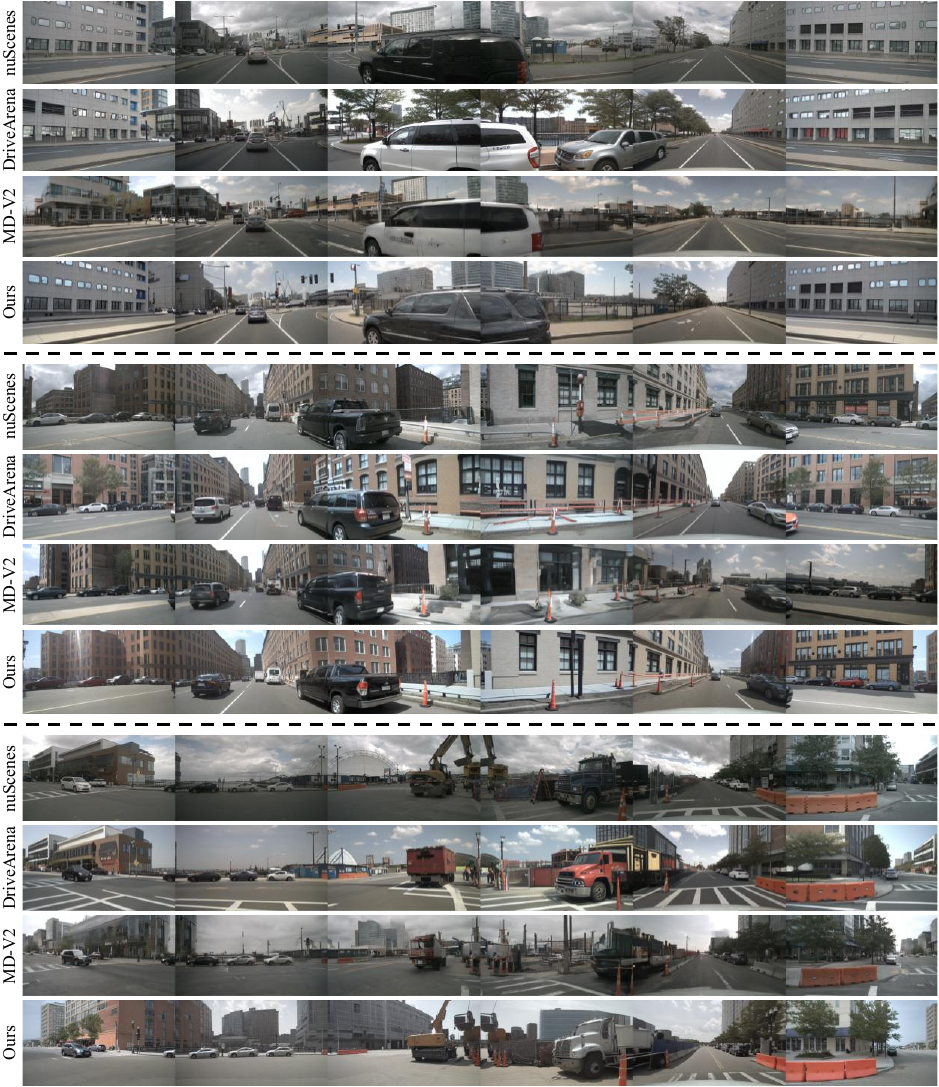}
    \vspace{-5pt}
    \caption{Qualitative comparison of generated samples. MD-V2 denotes MagicDrive-V2~\cite{gao2025magicdrive}.}
    \label{fig:more_qual_comp}
\end{figure*}

\Cref{fig:more_qual_comp} shows additional multi-view qualitative comparisons. Across diverse driving scenes, our method consistently preserves scene structure and maintains cross-view consistency, while the competing approach often removes scene elements or produces view-inconsistent object shapes.


\bibliographystyle{splncs04}
\bibliography{main}
\end{document}